\pdfoutput=1
\documentclass[runningheads]{llncs}
\usepackage{graphicx}
\usepackage{comment}
\usepackage{amsmath,amssymb} % define this before the line numbering.
\usepackage{color}

\usepackage{microtype}
\usepackage{booktabs}
\usepackage{multirow}
\usepackage{bbm}
\usepackage[caption=false, font=normalsize, labelfont=sf, textfont=sf]{subfig}
\usepackage[mathscr]{euscript}

\usepackage{enumitem}
\usepackage{blindtext}
\usepackage{makecell}

\usepackage{pbox}
\usepackage{caption}
\usepackage[pagebackref=true,breaklinks=true,letterpaper=true,colorlinks,bookmarks=false]{hyperref}
\makeatletter
\newcommand{\printfnsymbol}[1]{%
  \textsuperscript{\@fnsymbol{#1}}%
}
\makeatother

\begin{document}
\frenchspacing
\pagestyle{headings}
\mainmatter
\def\ECCVSubNumber{3742}  % Insert your submission number here

\title{VQA-LOL: Visual Question Answering under the Lens of Logic} % Replace with your title
\titlerunning{VQA-LOL}
\author{
Tejas Gokhale\thanks{Equal Contribution}\orcidID{0000-0002-5593-2804} \and
Pratyay Banerjee \printfnsymbol{1}\orcidID{0000-0001-5634-410X}  \and
Chitta Baral\orcidID{0000-0002-7549-723X} \and 
Yezhou Yang\orcidID{0000-0003-0126-8976}
}
\authorrunning{T. Gokhale et al.}
\institute{Arizona State University, United States\\
\email{\{tgokhale, pbanerj6, chitta, yz.yang\}@asu.edu}}

\maketitle

\begin{center}
    \includegraphics[width=\linewidth]{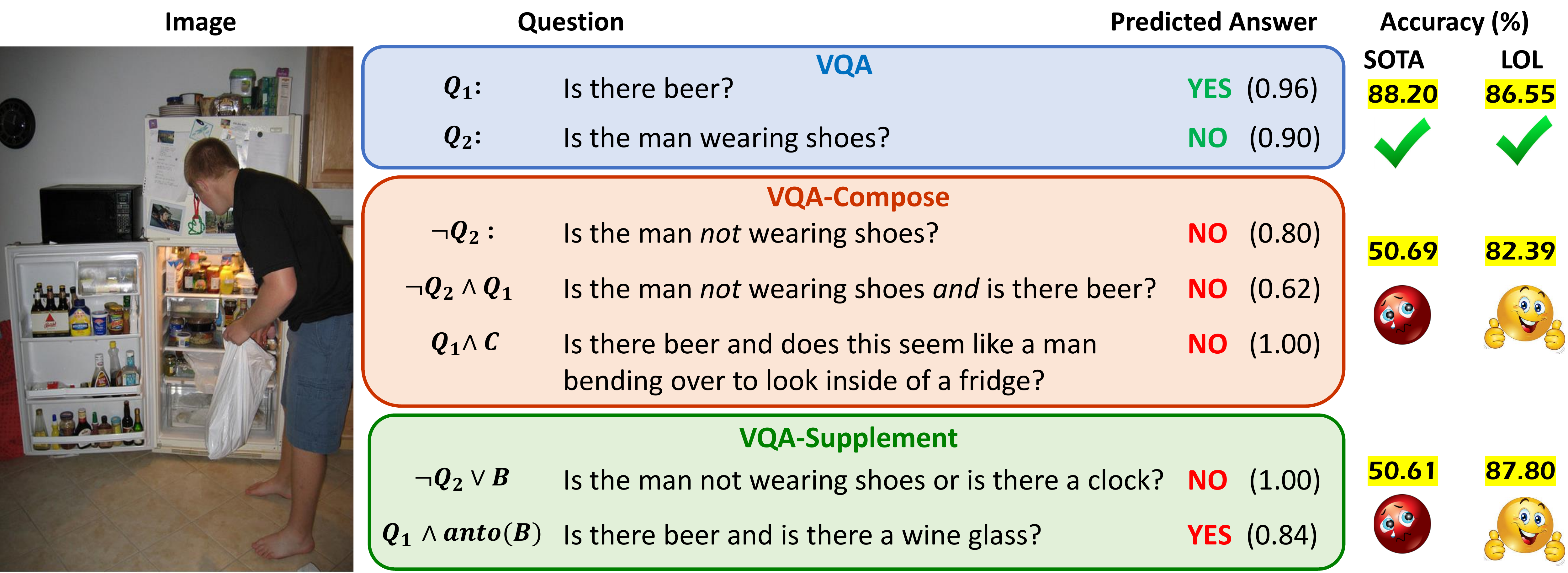}
    \captionof{figure}{
        % Illustration of logical composition of questions. 
        State-of-the-art models answer questions from the VQA dataset ($Q_1, Q_2$) correctly, but struggle when asked a logical composition including negation, conjunction, disjunction, and antonyms. 
        We develop a model that improves on this metric substantially, while retaining VQA performance.
        }
    \label{fig:motivation}
\end{center}%

%%%%%%%%%%%%%%%%%%%%%%%%%%%%%%%%%%%%%%%%%%%%%%%%%%%%%%%%%%%%%%%%
\begin{abstract}
%%%%%%%%%%%%%%%%%%%%%%%%%%%%%%%%%%%%%%%%%%%%%%%%%%%%%%%%%%%%%%%%
Logical connectives and their implications on the meaning of a natural language sentence are a fundamental aspect of understanding.
In this paper, we investigate whether visual question answering (VQA) systems trained to answer a question about an image, are able to answer the logical composition of multiple such questions.
When put under this \textit{Lens of Logic}, state-of-the-art VQA models have difficulty in correctly answering these logically composed questions.
We construct an augmentation of the VQA dataset as a benchmark, with questions containing logical compositions and linguistic transformations (negation, disjunction, conjunction, and antonyms).
We propose our {Lens of Logic (LOL)} model which uses question-attention and logic-attention to understand logical connectives in the question, and a novel Fréchet-Compatibility Loss, which ensures that the answers of the component questions and the composed question are consistent with the inferred logical operation.
Our model shows substantial improvement in learning logical compositions while retaining performance on VQA.
We suggest this work as a move towards robustness by embedding logical connectives in visual understanding.

\end{abstract}

%%%%%%%%%%%%%%%%%%%%%%%%%%%%%%%%%%%%%%%%%%%%%%%%%%%%%%%%%%%%%%%%
\section{Introduction}
%%%%%%%%%%%%%%%%%%%%%%%%%%%%%%%%%%%%%%%%%%%%%%%%%%%%%%%%%%%%%%%%
Theories about logic in human understanding have a long history. 
In modern times, Piaget and Fodor~\cite{piattelli1980language} studied the representation of logical hypotheses in the human mind.
George Boole~\cite{boole1854investigation} formalized conjunction, disjunction, and negation into an ``algebra of thought'' as a way to improve, systemize, and mathematize Aristotle's Logic~\cite{corcoran1972completeness}.
Horn regarded negation to be a fundamental and defining characteristic of human communication~\cite{horn2000negation}, following the traditions of Sankara~\cite{raju1954principle}, Spinoza~\cite{spinoza1934ethics}, and Hegel~\cite{hegel1929hegel}.
Recent studies~\cite{arlotti1263} have suggested that infants can formulate intuitive and stable logical structures to interpret dynamic scenes and to entertain and rationally modify hypotheses about the scenes.
As such we argue that understanding logical structures in questions, is a fundamental requirement for any question-answering system.
\begin{quotation}
    \begin{quote}
        \textit{If a question can be put at all, then it can be answered.}~\cite{wittgenstein2013tractatus}
    \end{quote}
\end{quotation}
In the above proposition, Wittgenstein linked the process of asking a question with the existence of an answer.
While we do not comment on the existence of an answer, we suggest the following softer proposition -
\begin{quotation}
    \begin{quote}
        \textit{If questions $Q_1\dots Q_n$ can be answered, then so should all composite questions created from $Q_1\dots Q_n$}
    \end{quote}
\end{quotation}

Visual question answering (VQA)~\cite{antol2015vqa} is an intuitive, yet challenging task that lies at a crucial intersection of vision and language. 
Given an image and a question about it, the goal of a VQA system is to provide a free-form or open-ended answer.
Consider the image in Figure~\ref{fig:motivation} which shows a person in front of an open fridge.
When asked the questions $Q_1$ ({\it Is there beer?}) and $Q_2$ ({\it Is the man wearing shoes?}) independently, the state-of-the-art model LXMERT~\cite{tan2019lxmert} answers both correctly.
However when we insert a negation in $Q_2$ ({\it Is the man not wearing shoes?}) or for a conjunction of two questions $\neg Q_2 \wedge Q1$ ({\it Is the man not wearing shoes and is there beer?}), the system makes wrong predictions. 
Our motivation is to reliably answer such logically composed questions.
In this paper, we analyze VQA systems under this {\it Lens of Logic (LOL)} and develop a model that can answer such questions reflecting human logical inference.
We offer our work as the first investigation into the logical structure of questions in visual question-answering and provide a solution that {\it learns} to interpret logical connectives in questions.

The first question is: can models pre-trained on the VQA dataset answer logically composed questions?
It turns out that these models are unable to do so, as illustrated in Figure~\ref{fig:motivation} and Table~\ref{table:exp1}.
An obvious next experiment is to \textit{split the question} into its component questions, predict the answer to each, and combine the answers logically. 
However language parsers (either oracle or trained parsers) are not accurate at understanding negation, and as such this approach does not yield correct answers for logically composed questions.
The question then arises: can the model answer such questions, if we explicitly train it with data that also contains logically composed questions?
For this investigation, we construct two datasets, \texttt{VQA-Compose} and \texttt{VQA-Supplement}, by utilizing annotations from the VQA dataset, as well as object and caption annotations from COCO~\cite{lin2014microsoft}.
We use these datasets to train the state-of-the-art model LXMERT~\cite{tan2019lxmert} and perform multiple experiments to test for robustness towards logically composed questions.

After this investigation, we develop our LOL model architecture that jointly learns to answer questions while understanding the type of question and which logical connective exists in the question, through our attention modules, as shown in Figure~\ref{fig:model}.
We further train our model with a novel Fréchet-Compatibility loss that ensures compatibility between the answers to the component questions and the answer of the logically composed question.
One key finding is that our models are better than existing models trained on logical questions, with a small deviation from state-of-the-art on VQA test set.
Our models also exhibit better {\it Compositional Generalization} i.e. models trained to answer questions with a single logical connective are able to answer those with multiple connectives.

Our contributions are summarized below:
\begin{enumerate}[noitemsep]
    \item We conduct a detailed analysis of the performance of the state-of-the-art VQA model with respect to logically composed questions,
    \item We curate two large scale datasets \texttt{VQA-Compose} and \texttt{VQA-Supplement} that contain logically composed binary questions.
    \item We propose \textit{LOL} -- our end-to-end model with dedicated attention modules that answer questions by understanding the logical connectives in questions.
    \item We show a capability of answering logically composed questions, while retaining VQA performance. % on VQA data.
\end{enumerate}

%%%%%%%%%%%%%%%%%%%%%%%%%%%%%%%%%%%%%%%%%%%%%%%%%%%%%%%%%%%%%%%%
\section{Related Work}
%%%%%%%%%%%%%%%%%%%%%%%%%%%%%%%%%%%%%%%%%%%%%%%%%%%%%%%%%%%%%%%%
\noindent\textbf{Logic in Human Expression:}
Is logical thinking a natural feature of human thought and expression?
Evidence in psychological studies~\cite{carey1985conceptual,gopnik1999scientist,arlotti1263} suggests that infants are capable of logical reasoning, toddlers understand logical operations in natural language and are able to compositionally compute meanings even in complex sentences containing multiple logical operators.
Children are also able to use these meanings to assign truth values to complex experimental tasks.
Given this, question-answering systems also need to answer compositional questions, and be robust to the manifestation of logical operators in natural language.\\

\noindent\textbf{Logic in Natural Language Understanding:}
The task of understanding compositionality in question-answering (QA) can also be interpreted as understanding logical connectives in text.
While question compositionality is largely unstudied, approaches in natural language understanding seek to transform sentences into symbolic formats such as first-order logic (FOL) or relational tables~\cite{mintz2009distant,zettlemoyer2012learning,lewis-steedman-2013-combined}.
While such methods benefit from interpretability, they suffer from practical limitations like intractability, reliance on background knowledge, and failure to process noise and uncertainty. 
\cite{bordes2013translating,riedel-etal-2013-relation,socher2013reasoning} suggest that better generalization can be achieved by learning embeddings to reason about semantic relations, and to simulate FOL behavior~\cite{rocktaschel2014low}.
Recursive neural networks have been shown to learn logical semantics on synthetic English-like sentences by using embeddings~\cite{bowman2014recursive,neelakantan2015compositional}.

Detection of negation in text has been studied for information extraction and sentiment analysis~\cite{morante2012modality}.
\cite{kassner2019negated} have shown that BERT-based models~\cite{devlin2018bert,liu2019roberta} are incapable of differentiating between sentences and their negations.
Concurrent to our work,~\cite{asai-hajishirzi-2020-logic} show the efficacy of FOL-guided data augmentation for performance improvements on natural language QA tasks that require reasoning.
Since our work deals with both vision and language modalities, it encounters a greater degree of ambiguity, thus calling for robust VQA systems that can deal with logical transformations.\\

\noindent\textbf{Visual Question Answering}
(VQA)~\cite{antol2015vqa} is a large-scale, human-annotated dataset for open-ended question-answering on images.
VQA-v2\cite{goyal2017making} reduces the language bias in the dataset by collecting complementary images for each question-image pair.
This ensures that the number of questions in the VQA dataset with the answer ``YES" is equal to those with the answer ``NO".
This dataset contains 204k images from MS-COCO~\cite{lin2014microsoft}, and 1.1M questions.

Cross-modal pre-trained models ~\cite{tan2019lxmert,lu2019vilbert,zhou2020unified} have proved to be highly effective in vision-and-language tasks such as VQA, referring expression comprehension, and image retrieval. 
While neuro-symbolic approaches~\cite{Mao2019NeuroSymbolic} have been proposed for VQA tasks which require reasoning on synthetic images, their performance on natural images is lacking.
Recent work seeks to incorporate reasoning in VQA, such as visual commonsense reasoning~\cite{zellers2019recognition,fang2020video2commonsense}, spatial reasoning~\cite{hudson2019gqa,johnson2017clevr}, and by integrating knowledge for end-to-end reasoning~\cite{Aditya:2019:IKR:3367722.3367926}.

We take a step back and extensively analyze the pivotal task of VQA with respect to various aspects of generalization.
We consider a rigorous investigation of a task, dataset, and models to be equally important as proposing new challenges that are arguably harder.
In this paper we analyse existing state-of-the-art VQA models with respect to their robustness to logical transformations of questions.

%%%%%%%%%%%%%%%%%%%%%%%%%%%%%%%%%%%%%%%%%%%%%%%%%%%%%%%%%%%%%%%%
\section{The Lens of Logic}
%%%%%%%%%%%%%%%%%%%%%%%%%%%%%%%%%%%%%%%%%%%%%%%%%%%%%%%%%%%%%%%%
\begin{table}[!t]
    \caption{Illustration of question composition in \texttt{VQA-Compose}, for the same example as in Figure \ref{fig:motivation}.
    QF: Question Formula, AF: Answer Formula}
    \begin{center}
    \resizebox{\textwidth}{!}{
    \begin{tabular}{@{}p{1.8cm}p{7.8cm}p{1.8cm}l@{}}
    \toprule 
    \textbf{QF} & \textbf{Question} & \textbf{AF} & \textbf{Answer}\\
    \midrule
    $Q_1$                       & Is there beer?                                    & $A_1$     & Yes \\
    $Q_2$                       & Is the man wearing shoes?                         & $A_2$     & No \\
    ${\neg}Q_1$                  & Is there no beer?                                 & ${\neg}A_1$ & No\\
    ${\neg}Q_2$                  & Is the man not wearing shoes?                     & ${\neg}A_2$ & Yes\\
    $Q_1{\wedge}Q_2$            & Is there beer and is the man wearing shoes?       & $A_1{\wedge}A_2$ & No \\
    $Q_1{\vee}Q_2$              & Is there beer or is the man wearing shoes?        & $A_1{\vee}A_2$ & Yes \\
    $Q_1{\wedge}{\neg}Q_2$       & Is there beer and is the man not wearing shoes?   & $A_1{\wedge}{\neg}A_2$ & Yes \\
    $Q_1{\vee}{\neg}Q_2$         & Is there beer or is the man not wearing shoes?    & $A_1{\vee}{\neg}A_2$ & Yes \\
    ${\neg}Q_1{\wedge}Q_2$       & Is there no beer and is the man wearing shoes?    & ${\neg}A_1{\wedge}A_2$ & No \\
    ${\neg}Q_1{\vee}Q_2$         & Is there no beer or is the man wearing shoes?     & ${\neg}A_1{\vee}A_2$ & No \\
    ${\neg}Q_1{\wedge}{\neg}Q_2$  & Is there no beer and is the man not wearing shoes?& ${\neg}A_1{\wedge}{\neg}A_2 $ & No \\
    ${\neg}Q_1{\vee}{\neg}Q_2$    & Is there no beer or is the man not wearing shoes? & ${\neg}A_1{\vee}{\neg}A_2$ & Yes \\
    \bottomrule
    \end{tabular}
    }
    \end{center}
    \label{tab:compose}
\end{table}

\begin{figure}[t]
    \centering
    \includegraphics[width=\linewidth]{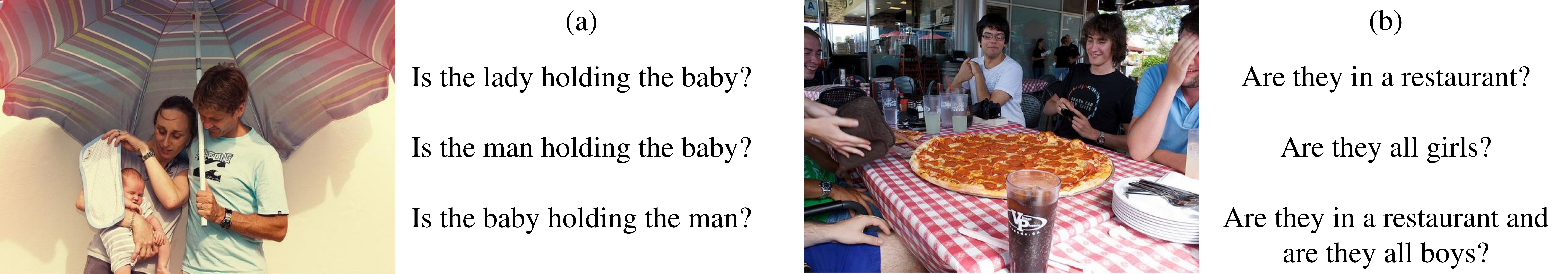}
    \caption{Some questions in \texttt{VQA-Supplement} created with adversarial antonyms.}
    \label{fig:illus}
\end{figure}

A lens magnifies objects under investigation, by allowing us to zoom and focus on desired contents or processes.
Our lens of logical composition of questions, allows us to magnify, identify, and analyze the problems in VQA models.

Consider Figure~\ref{fig:illus}(a), where we transform the first question {\it ``Is the lady holding the baby"} by first replacing {\it ``lady"} with an adversarial antonym {\it ``man"} and observe that the system provides a wrong answer with very high probability.
Swapping {\it ``man"} with {\it ``baby"} results in a wrong answer as well.
In~\ref{fig:illus}(b) a conjunction of two questions containing antonyms ({\it girls} vs {\it boys}) yields a wrong answer.
We identify that the ability to answer composite questions created by negation, conjunction and disjunction of questions is crucial for VQA.

We use ``closed questions" as defined in~\cite{bobrow1964natural} to construct logically composed questions. 
Under this definition, if a closed question has a negative (``NO") answer then its negation must have an affirmative (``YES") answer.
Of the three types of questions in the VQA dataset (yes/no, numeric, other), `yes-no" questions satisfy this requirement.
Although, visual questions in the VQA dataset can have multiple correct answers~\cite{bhattacharya2019does}, $20.91\%$ of the questions (around 160k) in the VQA dataset are closed questions, i.e. questions with a single unambiguous yes-or-no answer, unanimously annotated by multiple human workers.
This allows us to treat these questions as propositions and create a truth table for answers to compose logical questions as shown in Table~\ref{tab:compose}.

    \subsection{Composite Questions}
    Let $\mathcal{D}$ be the VQA dataset. 
    For closed questions $Q_1$ and $Q_2$ about image $I\in\mathcal{D}$, we define the composite question $Q^*$ composed using connective $\circ\in\{ \vee, \wedge\}$, as:
    \begin{equation}
        Q^* = \widehat{Q_1} ~\circ ~\widehat{Q_2}, \qquad where~~ \widehat{Q_1} \in \{ Q_1, \neg Q_1\},  ~ \widehat{Q_2} \in \{ Q_2, \neg Q_2\}. 
    \end{equation}
    
    \subsection{Dataset Creation Process}
    Using the above definition we create two new datasets by utilizing multiple questions about the same image (\texttt{VQA-Compose}) and external object and caption annotations about the image from COCO to create more questions (\texttt{VQA-Supplement}).
    The seed questions for creating these datasets are all closed binary questions from VQA-v2~\cite{goyal2017making}.
    These datasets serve as test-beds, and enable experiments that analyze performance of models when answering such questions.\\

        \noindent\textbf{VQA-Compose:}
        Consider the first two rows in Table~\ref{tab:compose}. 
        $Q_1$ and $Q_2$ are two questions about the image in Figure~\ref{fig:motivation} taken from the VQA dataset.
        Additional questions are composed from $Q_1$ and $Q_2$ by using the formulas in Table~\ref{tab:compose}.
        Thus for each pair of closed questions in the VQA dataset, we get 10 logically composed questions.
        Using the same train-val-test split as the VQA-v2 dataset~\cite{goyal2017making}, we get \textit{1.25 million samples} for our \texttt{VQA-Compose} dataset.
        The dataset is balanced in terms of the number of questions with affirmative and negative answers.\\
        
        \noindent\textbf{VQA-Supplement:}
        Images in VQA-v2 follow identical train-val-test splits as their source MS-COCO~\cite{lin2014microsoft}.
        Therefore, we use the object annotations from COCO to create additional closed binary questions, such as {\it ``Is there a bottle"} for the example in Figure~\ref{fig:motivation}.
        We also create ``adversarial" questions about objects, like {\it ``Is there a wine-glass?"} by using an object that is not present in the image (wine-glass), but is \textit{semantically close} to an object in the image (bottle).
        We use Glove vectors~\cite{pennington2014glove} to find the adversarial object with the closest embedding. 
        Following a similar strategy, we also convert captions provided in COCO to closed binary questions, for example {\it ``Does this seem like a man bending over to look inside the fridge"}.
        Since we know what objects are present in the image, and the captions describe a ``true" scene, we are able to obtain the ground-truth answers for questions created from objects and captions.
        Similar methods for creation of question-answer pairs have previously been used in~\cite{ren2015exploring,malinowski2014multi}.

        Thus for every question, we obtain several questions from objects and captions, and use these to compose additional questions by following a process similar to the one for \texttt{VQA-Compose}.
        For each closed question in the VQA dataset, we get $20$ additional logically composed questions by utilizing questions created from objects and captions, yielding a total of \textit{2.55 million samples} as \texttt{VQA-Supplement}.

    \subsection{Analytical Setup}
    In order to test the robustness of our models to logically composed questions, we devise five key experiments to analyse baseline models and our methods.
    These experiments help us gain insights into the nuances of the VQA dataset, and allow us to develop strategies for promoting robustness.\\

        \noindent\textbf{Effect of Data Augmentation:}
        In this experiment, we compare the performance of models on \texttt{VQA-Compose} and \texttt{VQA-Supplement} with or without logically composed training data.
        This experiment allows us to test our hypotheses about the robustness of any VQA model to logically composed questions.
        We first use models trained on VQA data to answer questions in our new datasets and record performance.
        We then explicitly train the same models with our new datasets, and make a comparison of performance with the pre-trained baseline.\\
        
        \noindent\textbf{Learning Curve:}
        We train our models with an increasing number of logically composed questions and compare performance.
        This serves as an analysis of the number of logical samples needed by the model to understand logic in questions.\\
    
        \noindent\textbf{Training only with Closed Questions:}
        In this ablation study, we restrict the training data to only closed questions i.e. ``Yes-No" VQA questions, \texttt{VQA-Compose} and \texttt{VQA-Supplement}, allowing our model to focus solely on closed questions.\\
    
        \noindent\textbf{Compositional Generalization:}
        We address whether training on closed questions containing single logical operation ($\neg Q_1$,  $Q_1\vee Q_2$) can generalize to multiple operations ($Q_1 \wedge \neg Q_2$, $\neg Q_1 \vee Q_2$).
        For instance, rows 1 through 6 in Table~\ref{tab:compose} are {\it single operation questions}, while rows 7 through 12 are {\it multi-operation questions}.
        Our aim is to have models that exhibit such compositional generalization.\\
        
        \noindent\textbf{Inductive Generalization:}
        We investigate if training on compositions of two questions ($\neg Q_1 \vee Q_2$) can generalize to compositions of more than two questions ($Q_1 \wedge \neg Q_2 \wedge Q_3 \dots$).
        This studies whether our models develop an understanding of logical connectives, as opposed to simply learning patterns from large data.

%%%%%%%%%%%%%%%%%%%%%%%%%%%%%%%%%%%%%%%%%%%%%%%%%%%%%%%%%%%%%%%%
\section{Method}
%%%%%%%%%%%%%%%%%%%%%%%%%%%%%%%%%%%%%%%%%%%%%%%%%%%%%%%%%%%%%%%%
\begin{figure}[t]
    \centering
    \includegraphics[width=\linewidth]{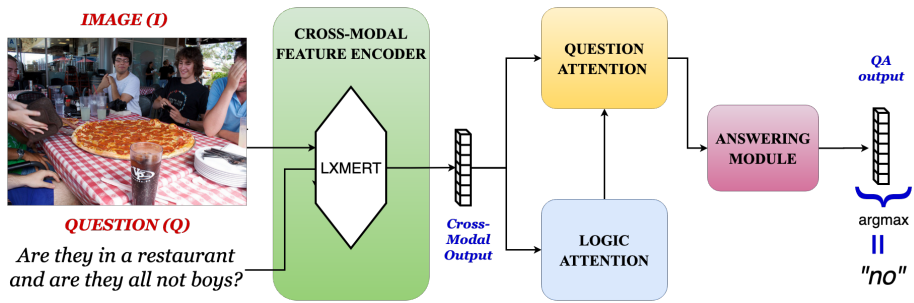}
    \caption{LOL model architecture showing a cross-modal feature encoder followed by our Question-Attention ($q_{\textit{\tiny ATT}}$) and Logic Attention ($\ell_{\textit{\tiny ATT}}$) modules. 
    The concatenated output of  is used by the Answering Module to predict the answer.
    % to the question (Q).
    }
    \label{fig:model}
\end{figure}

In this section. we describe LXMERT~\cite{tan2019lxmert} (a state-of-the-art VQA model), our Lens of Logic (LOL) model, attention modules which learn the question-type and logical connectives in the question, and the Fréchet-Compatibility (FC) Loss.
This section refers to a composition of two questions, but applies to $n\geq2$ questions.

    \subsection{Cross-Modal Feature Encoder}
    LXMERT (Learning Cross-Modality Encoder Representations from Transformers)~\cite{tan2019lxmert} is one of the first cross-modal pre-trained frameworks for vision-and-language tasks, that combines a strong visual feature extractor~\cite{ren2015faster} with a strong language model (BERT)\cite{devlin2018bert}.
    LXMERT is pre-trained for key vision-and-language tasks, on a large corpus of $\sim$9M image-sentence pairs, making it a powerful cross-modal encoder for vision+language tasks such as visual question answering, as compared to other models such as MCAN~\cite{Yu_2019_CVPR} and UpDn~\cite{Anderson2017up-down}, and strong representative baseline for our experiments.

    \subsection{Our Model: Lens of Logic (LOL)}
    The design for our LOL model is driven by three key insights:
    \begin{enumerate}[noitemsep]
        \item As logically composed questions are closed questions, understanding the type of question will guide the model to answer them correctly.
        \item Predicted answers must be compatible with the predicted question type. For instance, a closed question can have an answer that is either ``Yes" or ``No".
        \item The model must learn to identify the logical connectives in a question.
    \end{enumerate}

    \noindent Given these insights, we develop the Question Attention module that encodes the type of question (\textit{Yes-No}, \textit{Number}, or \textit{Other}), and the Logic Attention module that predicts the connectives (\textit{AND, OR, NOT, no connective}) present in the question, and use these to learn representations.
    The overall model architecture is shown in Figure~\ref{fig:model}.
    For every question $Q$ and corresponding image $I$, we obtain embeddings $z_Q$ and $z_I$ respectively, as well as a cross-modal embedding $z_X$. 

        \textbf{Question Attention Module ($q_{\textit{ATT}}$)}
        takes cross-modal embedding $z_x$ from LXMERT as input, and outputs vector \textbf{$P_{type} = softmax(\mathbf{q_{\textit{\tiny ATT}}}(z_x))$}, representing the probabilities of each question-type. 
        These probabilities are used to get a final representation $\mathbf{z^{type}}$ which combines the features for each question-type.\footnotemark[1]

        \textbf{Logic Attention Module ($\ell_{\textit{ATT}}$)}
        takes the cross-modal embedding $z_X$ from LXMERT as input, and outputs vector \textbf{$P^{conn} = \sigma(\mathbf{\ell_{\textit{ATT}}}(z_X))$} which represents the probabilities of each type of connective.
        We use sigmoid ($\sigma$) instead of a softmax, since a question can have multiple connectives.
        These probabilities are used to combine the features for each type of connective into a final representation $\mathbf{z^{conn}}$ which encodes information about the connectives in the question.

    \subsection{Loss Functions}
    We train our models jointly with the loss function given by:
    \begin{equation}
        \mathcal{L} = (1{-}\alpha_1{-}\alpha_2)\cdot \mathcal{L}_{ans} + \alpha_1 \cdot \mathcal{L}_{type} + \alpha_2 \cdot \mathcal{L}_{conn} + \beta\cdot\mathcal{L}_{\mathit{FC}}.
    \end{equation}

        \noindent\textbf{Answering Loss} $\ell_{ans}$ is conditioned on the type of question.
        We multiply the final prediction vector with the probability and the mask $M_i$ for question-type $i$.
        $M_i$ is a binary vector with $1$ for every answer-index of type-i and $0$ elsewhere:
        
        \begin{equation}
            \mathcal{L}_{\textit{ans}} = \mathcal{L}_{\textit{\tiny BCE}}(\sum_{i=1}^{3}\hat{y} \odot M_i \cdot P^{\textit{type}}_i, y_{\textit{ans}}).
        \end{equation}
        
        \noindent\textbf{Attention Losses:}
        $q_{\textit{\tiny ATT}}$ is trained to minimize a Negative Log Likelihood (NLL) classification loss, ensuring a shrinkage of probabilities of the answer choices of the wrong type. $\ell_{\textit{\tiny ATT}}$ is trained to minimize a multi-label classification loss, using Binary Cross-Entropy (BCE) given by:
        \begin{align}
            \mathcal{L}_{\textit{type}} &= \mathcal{L}_{\textit{\tiny NLL}}(\text{softmax}(z^{\textit{type}}), y_{\textit{type}}),\\
             \mathcal{L}_{\textit{conn}} &= \mathcal{L}_{\textit{\tiny BCE}}(\sigma(z^{\textit{conn}}), y_{\textit{conn}}),
        \end{align}
        where $y_{ans}, y_{type}, y_{conn}$ are labels for answer, question-type and connective.\\

        \noindent\textbf{Fréchet-Compatibility Loss:}
        We introduce a new loss function that ensures compatibility between the answers predicted by the model for the component questions $Q_1$ and $Q_2$ and the composed question $Q$.
        Let $A, A_1, A_2$ be the respective answers predicted by the model for $Q$, $Q_1$, and $Q_2$. $Q_i$ can have negation.
        Then Fréchet inequalities~\cite{boole1854investigation,frechet1935generalisation} provide us with bounds for the probabilities of the answers of the conjunction and disjunction of the two questions:
        \begin{align}
            \mathit{max}(0, p(A_1)+p(A_2)-1) &\leq p(A_1 \wedge A_2) \leq \mathit{min}(p(A_1), p(A_2)). \\
            \mathit{max}(p(A_1), p(A_2)) &\leq p(A_1 \vee A_2) \leq \mathit{min}(1, p(A_1) + p(A_2)).
        \end{align}
        We define ``Fréchet bounds" $b_L$ and $b_R$ to be the left and right bounds for the triplet $A, A_1, A_2$,
        and the ``Fréchet Mean" $m_A$ to be the average of the Fréchet bounds; $m_A = (b_L + b_R)/2$.
        Then, the Fréchet-Compatibility Loss given by:
        \begin{equation}
            \mathcal{L}_{FC} = (p(A) - \mathbbm{1}(m_A > 0.5))^2, 
        \end{equation}
        ensures that the predicted answer and that determined by $m_A$ match. 

    \subsection{Implementation Details}
    The LXMERT feature encoder produces a vector $z$ of length 768 which is used by our attention modules, each having sub-networks $\mathbf{f_i}, \mathbf{g_i}$ with 2 feed-forward layers.
    We first train our models without FC loss.
    Then we select the best models with a checkpoint of 10 epochs and finetune these further for 3 epochs with FC loss, since the FC loss is designed to work for a model whose predictions are not random.
    Thus our improvements in accuracy are attributable to the FC Loss and not more training epochs.
    We utilize the Adam optimizer~\cite{kingma2014adam} with a learning rate of $5\textit{e-}5$, batch size of 32 and train for 20 epochs.
    Our models are trained on 4 NVIDIA V100 GPUs, and take approximately 24 hours for training 20 epochs.\footnote{More training details in Supplementary Materials}

%%%%%%%%%%%%%%%%%%%%%%%%%%%%%%%%%%%%%%%%%%%%%%%%%%%%%%%%%%%%%%%%%%%%%%%%%%%%%%%%
\section{Experiments}
%%%%%%%%%%%%%%%%%%%%%%%%%%%%%%%%%%%%%%%%%%%%%%%%%%%%%%%%%%%%%%%%%%%%%%%%%%%%%%%%
\begin{table}[t]
    \caption[Comparison of LXMERT and LOL when trained on VQA dataset and the combinations of VQA and \texttt{VQA-Compose}, \texttt{VQA-Supplement}.]{Comparison of LXMERT and LOL trained on VQA data, combinations with \texttt{Compose}, \texttt{Supplement}, and our Frechet-Compatibility (FC) Loss \footnotemark}
    \begin{center}
    \resizebox{0.85\linewidth}{!}{
    \begin{tabular}{p{2.5cm} p{3.5cm} p{1.2cm}p{1.2cm}p{1.2cm}p{1.2cm}}
    \toprule
    & & \multicolumn{4}{c}{\textbf{Validation Accuracy (\%) $\uparrow$}}\\
    \cmidrule{3-6}
    \textbf{Model} & \textbf{Trained on} & VQA & YN & Comp & Supp \\
    \toprule
    \multirow{3}{*}{\footnotesize LXMERT}
    & VQA &               68.94 & \textbf{86.65} & 50.79 & 50.51 \\
    & VQA + Comp &        67.85 & 85.32 & 85.03 & 80.85 \\
    & VQA + Comp + Supp & 68.83 & 84.83 & 70.28 & 85.17 \\

    \textit{ \footnotesize with FC Loss} & VQA + Comp + Supp & 67.84 &	84.92 & 75.31 &	85.25 \\
    \midrule
    \multirow{3}{*}{{\footnotesize LOL (qATT)}}
    & VQA &               \underline{\textbf{69.08}} & \underline{85.32} & 48.99 & 50.54 \\
    & VQA + Comp &        67.51 & 84.82 & 84.85 & 79.62 \\
    & VQA + Comp + Supp & 68.72 & 84.99 & 79.88 & 87.12 \\
    \midrule
    
    \multirow{3}{*}{\pbox{25mm}{\footnotesize LOL (Full)}}
    & VQA + Comp &        68.94 & 85.15 & \underline{\textbf{85.13}} & 79.02\\
    & VQA + Comp + Supp & 68.86 & 84.87 & 81.07 & 87.54\\

    \textit{ \footnotesize with FC Loss} & VQA + Comp + Supp & 68.10 &	84.75 & 82.39 &	\underline{\textbf{87.80} }\\
    
    \midrule 
    \midrule
    %% YES-NO only training
    \multirow{2}{*}{{\footnotesize LXMERT}}
    & YN + Comp & - & 84.13 & 84.44 & 79.39 \\
    & YN + Comp + Supp & - & 84.09 & 82.63 & 88.15 \\ 
    \midrule 
    
    %% YES-NO only training
    \multirow{2}{*}{{\footnotesize LOL ($\ell$ATT})}
    & YN + Comp & - & 85.22 & \underline{\textbf{85.31}} & 79.87 \\
    & YN + Comp + Supp & - & 85.26 & 84.37 & \underline{\textbf{89.00}}\\ 
    \bottomrule
    \end{tabular}
    }
    \end{center}
    
    \label{table:exp1}
\end{table}
\footnotetext{In all tables, best overall scores are bold, our best scores underlined.}
We first conduct analytical experiments to test for logical robustness and transfer learning capability.
We use three datasets for our experiment: the VQA v2.0~\cite{antol2015vqa} dataset, a combination of VQA and our \texttt{VQA-Compose} dataset, and a combination of VQA, \texttt{VQA-Compose} and \texttt{VQA-Supplement}.
The size of the training dataset and the distribution of yes-no, number and other questions is kept the same as the original VQA dataset ($\sim$443k) for fair comparison.
Since \texttt{VQA-Supplement} uses captions and objects from MS-COCO, we use is to analyze the ability of our models to generalize to a new source of data (MS-COCO) as well as questions containing adversarial objects.
After training, our attention modules ($q_{\textit{ATT}}$ and $\ell_{\textit{ATT}}$) achieve an accuracy of 99.9\% on average, showing almost perfect performance when it comes to learning the type of question and the logical connectives present in the question.

\begin{table}[t]
    
    \caption[Generalization]{Validation accuracies ($\%$) for Compositional Generalization and Commutative Property. Note that 50\% is random performance.\footnotemark[2]}
    \begin{center}
    \resizebox{\linewidth}{!}{
    \begin{tabular}{p{2cm} c c cc c cc}
    \toprule
    \multirow{2}{*}{\textbf{Model}} &  & \hphantom & \multicolumn{2}{c}{\textbf{VQA-Compose}} & \hphantom &  \multicolumn{2}{c}{\textbf{VQA-Supplement}}\\
    \cmidrule{4-5} \cmidrule{7-8}
     & \textbf{YN} & & Single & Multiple & & Single & Multiple\\
    \toprule
    \multirow{1}{*}{\pbox{10mm}{\footnotesize LXMERT}}
    % & YN + C          & 84.16 & 84.71  & 65.60 & 80.63 & 66.55 \\
    & 85.07 & & 83.95 & 61.99 & & 86.65 & 60.00 \\
    \midrule
    \multirow{1}{*}{\footnotesize LOL}
    % & YN + C & \underline{\textbf{85.13}} & \underline{\textbf{85.85}} & \underline{\textbf{66.87}} & 81.66 & \underline{\textbf{79.10}}\\
    & 85.12 & & 84.60 & 66.03 & & \underline{\textbf{87.42}} & 66.05  \\
    \bottomrule
    % \multirow{2}{*}{\pbox{15mm}{\footnotesize LXMERT + \\Type + Conn}}
    % & YN + C & 85.17 & 81.88 & \textbf{70.58} & 90.76 & 46.19 \\
    % & YN + C + S & 85.19 & 81.16 & 57.76 & 91.88 & 47.76  \\
    % \hline
    \end{tabular}

    \quad

     \begin{tabular}{p{2cm} cc c cc}
    \toprule
    \multirow{2}{*}{\textbf{Model}} & \multicolumn{2}{c}{\textbf{VQA-Compose}} & \hphantom &  \multicolumn{2}{c}{\textbf{VQA-Supplement}}\\
    \cmidrule{2-3} \cmidrule{5-6}
     & $Q_1\circ~Q_2$ & $Q_2\circ~Q_1$ & & $Q_1\circ~Q_2$ & $Q_2\circ~Q_1$\\
    \toprule
    \multirow{1}{*}{\pbox{10mm}{\footnotesize LXMERT}} & 
    82.34 & 80.44 & & 85.57 & 81.78\\
    \midrule
    \multirow{1}{*}{\footnotesize LOL} & 84.91 & 83.64 & & 85.62 & 83.41\\
    \bottomrule
    % \multirow{2}{*}{\pbox{15mm}{\footnotesize LXMERT + \\Type + Conn}}
    % & YN + C & 85.17 & 81.88 & \textbf{70.58} & 90.76 & 46.19 \\
    % & YN + C + S & 85.19 & 81.16 & 57.76 & 91.88 & 47.76  \\
    % \hline
    \end{tabular}
    }
    \end{center}
    
    \label{table:exp4}
\end{table}
\begin{figure}[t]
    \begin{center}
    \includegraphics[width=\linewidth]{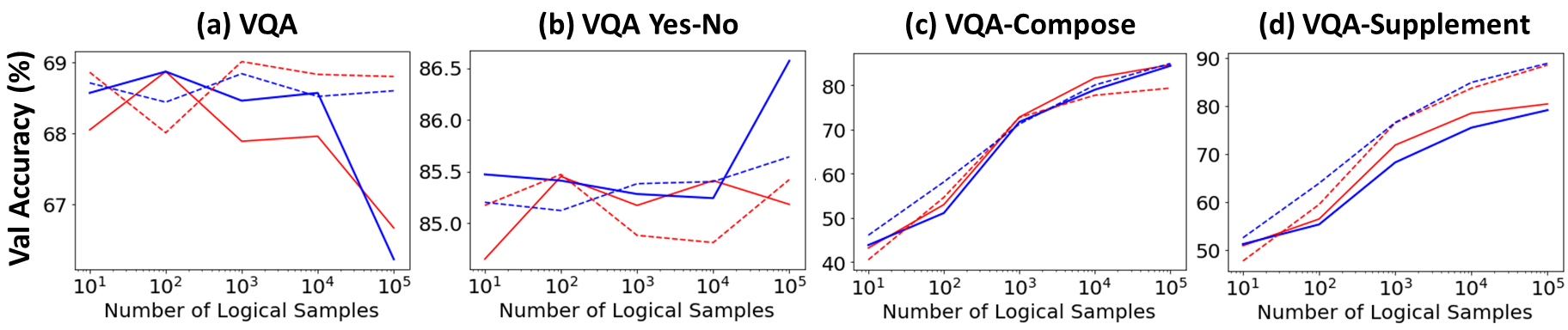}
    \caption{
    % Learning Curves: Red lines denote LXMERT, Blue lines denote our LOL model with $q_{\textit{ATT}}$. 
    % Models trained on VQA + \texttt{Comp} are shown in solid and those trained on VQA + \texttt{Comp} + \texttt{Supp} are in dotted lines. Best viewed in color.
    Learning Curve comparison for models (Red: LXMERT, Blue: LOL) trained on our datasets (solid lines: VQA~+~\texttt{Comp}, dotted lines: VQA~+~\texttt{Comp}~+~\texttt{Supp})
    }
    
    \label{fig:exp3}
    \end{center}
\end{figure}

    \subsection{Can't We Just Parse the Question into Components?}
    
    Since our questions are a composition of multiple questions, an obvious approach is to split the question into its components, and to discern the logical formula for composition.
    The answers to these component questions (predicted by VQA models) can be \textit{re-combined} with the predicted logical formula to obtain the final answer.
    We use parsers to map components and logical operations to predefined slots in a logical function.
    The oracle parser uses the ground truth component questions and combines predicted answers using the true formula.
    However, at test time we do not have access to the true mapping and components.
    So we train a RoBERTa-Base~\cite{liu2019roberta} parser using B-I-O tagging~\cite{ramshaw-marcus-1995-text} for a Named-Entity Recognition task with constituent questions as entities.\footnotemark[1]
    
    The performance of the oracle parser serves as the upper bound as we have a perfect mapping, with the QA system being the only source of error.
    The trained parser has an exact-match accuracy of $85\%$, but only a $72\%$ accuracy in determining the number of operands.
    The parser has an accuracy of $89\%$ for questions with 3 or less operands, but only $78\%$ for longer compositions.
    End-to-end (E2E) models do not need to parse questions and hence overcome these hurdles, but do require an understanding of logical operations.
    Table~\ref{table:sota} shows that both oracle and trained parsers when used with LOL outperform parsers with LXMERT, by $6.82\%$) and $5.60\%$ respectively.
    The LOL model without using any parsers is better than both LXMERT and LOL with the trained parser by $7.55\%$ and $1.95\%$ respectively.

    \subsection{Explicit Training with Logically Composed Questions}
    \noindent\textbf{Can models trained on the VQA-v2 dataset answer logically composed questions?}
    The first section of Table~\ref{table:exp1} shows that LXMERT, when trained only on questions from VQA-v2 has near random accuracy ($\sim$50\%) on our logically composed datasets, thus exhibiting little robustness to such questions.\\
    
    \noindent\textbf{Can baseline model improve if trained explicitly with logically composed questions questions?}
    We train the models with data containing a combination of samples from VQA-v2, \texttt{VQA-Compose}, and \texttt{VQA-Supplement}.
    The accuracy on \texttt{VQA-Compose} and \texttt{VQA-Supplement} 
    improves, but there is a drop in performance on yes-no questions from VQA.
    Our models with our attention modules ($q_{\textit{\tiny ATT}}$ and $\ell_{\textit{\tiny ATT}}$) are able to retain performance on VQA-v2 while achieving improvements on all validation datasets.

    \subsection{Analysis}

        \noindent\textbf{Training with Closed Questions only:}
        We analyse the performance of models when trained only with closed questions from VQA, VQA + \texttt{Comp} and VQA + \texttt{Comp} + \texttt{Supp} and see that our model achieves the best accuracy on logically composed questions, as shown in sections 3 and 4 in Table~\ref{table:exp1}.
        Since we train only closed questions, we do not use our question attention module for this experiment.\\
        
        \noindent\textbf{Effect of Logically Composed Questions:}
        We increase the number of logical samples in the training data on a log scale from 10 to 100k.
        As can be seen from the learning curves in Figure~\ref{fig:exp3}(a), models trained on VQA + \texttt{Comp} + \texttt{Supp} are able to retain performance on VQA validation data, while those trained only on VQA + \texttt{Comp} data deteriorate.
        Figure~\ref{fig:exp3}(b) shows that our models improve on VQA Yes-No performance after being trained on more logically composed samples, exhibiting transfer learning capabilities.
        In (c) both our models are comparable to the baseline, but our model shows improvements over the baseline when trained on VQA + \texttt{Comp} + \texttt{Supp}.
        In (d) for all levels of additional logical questions, our model trained on VQA + \texttt{Comp} + \texttt{Supp} is the best performing.
        From (c) and (d), we observe that a large number of logical questions are needed during training for the models to learn to answer them during inference.
        We also see that our model yields the best performance on \texttt{VQA-Supplement}.\\
        
        \noindent\textbf{Compositional Generalization:}
        To test for compositional generalization, we train models on questions with a maximum of one connective (single) and test on those with multiple connectives.
        It can be seen from Table~\ref{table:exp4} that our models are better equipped than the baseline to generalize to multiple connectives and also to be able to generalize from \texttt{VQA-Compose} to \texttt{Supplement}.\\
        
            \begin{figure}[t]
        \begin{center}
            \subfloat[\label{heatmap_comp}]{%
            \includegraphics[angle=0,origin=c,width=0.34\linewidth]{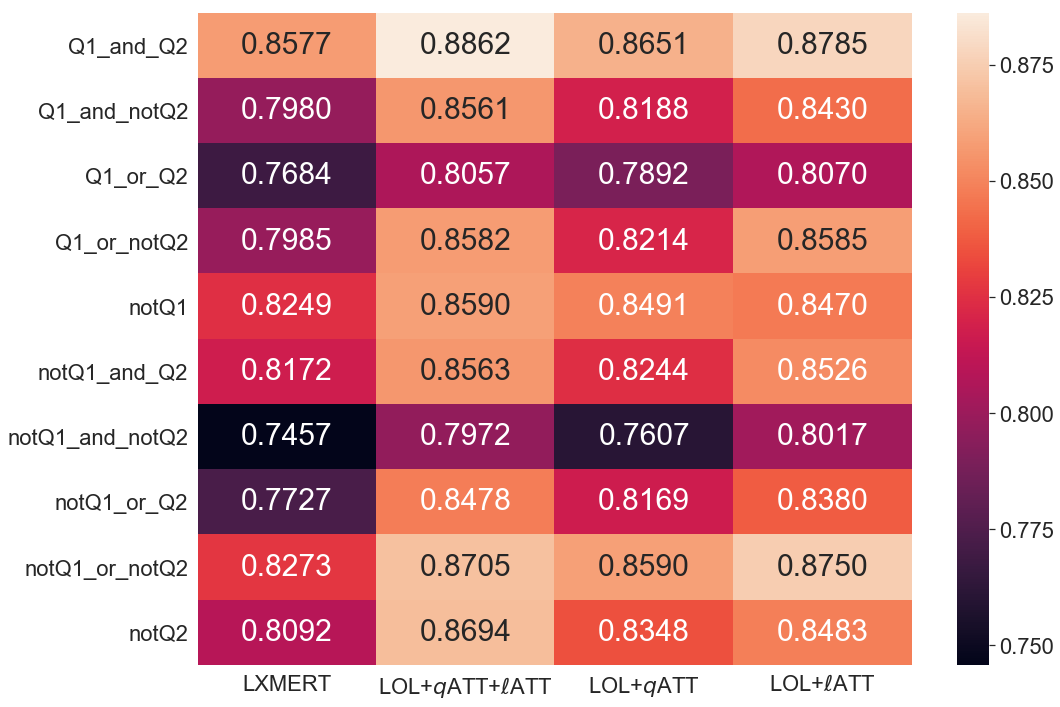}
            }
            \subfloat[\label{heatmap_supp}]{%
            \includegraphics[angle=0,origin=c,width=0.34\linewidth]{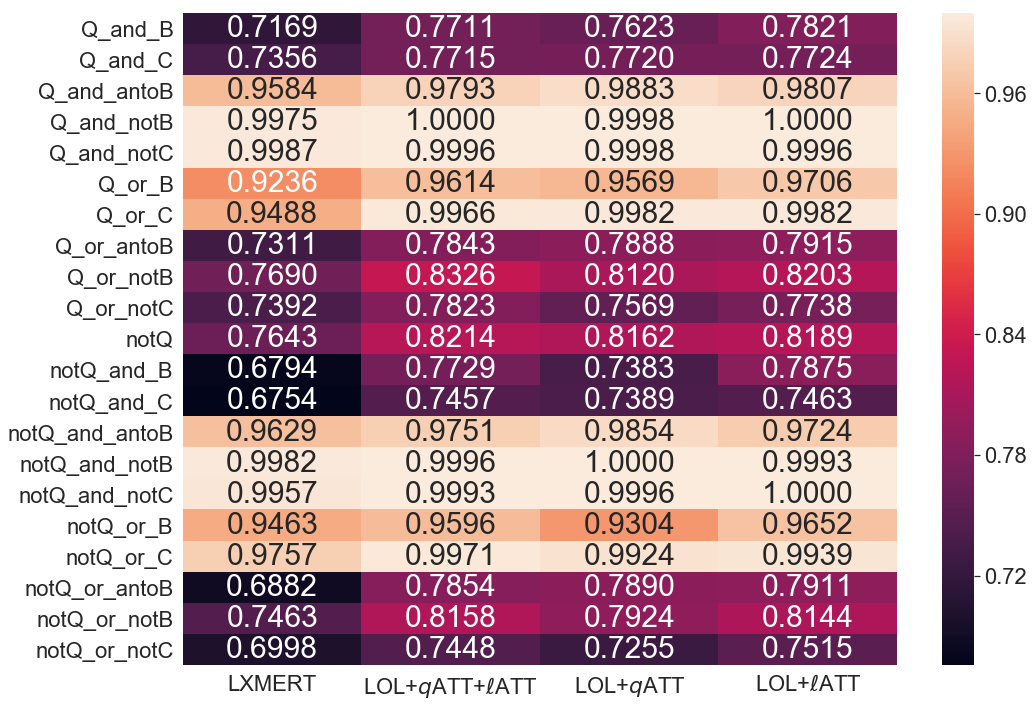}
            }
            \subfloat[\label{varlen}]{%
            \includegraphics[angle=0,origin=c,width=0.3\linewidth]{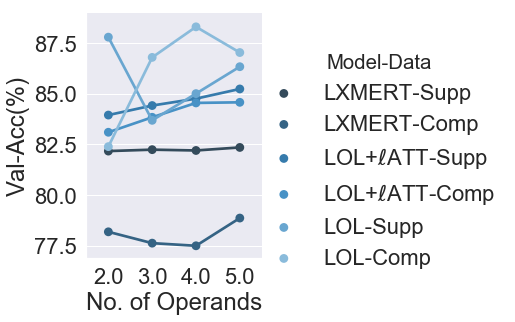}
            }
        \end{center}
        \caption{Accuracy for each type of question in (a) VQA-Compose, (b) VQA-Supplement and for questions with number of operands greater than 2.}
        \label{fig:heatmap}
    \end{figure}
        
        \noindent\textbf{Inductive Generalization:}
        We test our models on questions composed with more than two components. Parser-based  models have this property by default.
        As shown by Figure~\ref{varlen} our E2E models outperform the baseline LXMERT.\\

        \noindent\textbf{Commutative Property:}
        Our models have identical answers when the question is composed either as $Q_1\circ Q_2$ or $Q_2\circ Q_1$, for logical operation $\circ$, as shown in Table~\ref{table:exp4}.
        The parser-based models are agnostic to the order of components if the parsing is accurate, while our E2E models are robust to the order.\\

        \noindent\textbf{Accuracy per Category of Question Composition:}
        In Figure~\ref{fig:heatmap} we show a plot of accuracy versus question type for each model.
        $Q, Q_1, Q_2$ are questions from VQA, $B, C$ are object-based and caption-based questions from COCO respectively.
        From the results, we interpret that questions such as $Q\wedge antonym(B), Q\wedge\neg B, Q\wedge\neg C$ are easy because the model is able to understand absence of objects, therefore can always answer these questions with a ``NO".
        Similarly, $ Q\vee B, Q\vee C$ are easily answered since presence of the object makes the answer always ``YES".
        By simply understanding object presence many such questions can be answered. Figure~\ref{fig:heatmap} shows the model has the same accuracy for logically equivalent operations.

    \begin{table*}[t]
    \caption[Test Set Results]{Performance on `test-standard' set of VQA-v2 and validation set of our datasets. LOL performance is close to SOTA on VQA-v2, but significantly better at logical robustness. $^*$\text{MCAN} uses a fixed vocabulary 
    % so we are unable to evaluate on 
    that prohibits evaluation on \texttt{VQA-Supplement} which has questions created from COCO captions. 
    $^{\#}$Test-dev scores, since MCAN does not report test-std single-model scores\footnotemark[2]}
    \begin{center}
     \resizebox{\linewidth}{!}{
    \begin{tabular}{@{}lllcccccccc@{}}
    \toprule
    \multirow{2}{*}{Model} & \multirow{2}{*}{Parser} & \multirow{2}{*}{\textbf{\pbox{20mm}{Training \\Data}}} & \multicolumn{4}{c}{\textbf{Test-Std. Accuracy (\%) $\uparrow$}} & \hphantom & \multicolumn{3}{c}{\textbf{Val. Accuracy (\%) $\uparrow$}}\\
     \cmidrule{4-7} \cmidrule{9-11}
     & & & Yes-No & Number & Other & Overall & & {Compose} & Supplement & Overall \\
    \midrule
    MCAN & None & VQA~\cite{Yu_2019_CVPR} & $86.82^{\#}$ & $53.26^{\#}$ & $60.72^{\#}$ & 70.90 & & 52.42 & {*} & {*}\\
    LXMERT & None & VQA~\cite{tan2019lxmert} & \textbf{88.20} & \textbf{54.20} & \textbf{63.10} & \textbf{72.50} & & 50.79 & 50.51 & 50.65\\
    LOL ($\mathit{q}$ATT) & None & VQA &  \underline{87.33} & \underline{54.03} & \underline{62.40} & \underline{72.03} & & 48.99 & 50.54 & 49.77\\
    \midrule
    LXMERT & Oracle & VQA & 88.20 & 54.20 & 63.10 & 72.50 & & 86.38 & 74.29 & 80.33 \\
    LXMERT & Trained & VQA & 88.20 & 54.20 & 63.10 & 72.50 & & 86.35 & 68.75 & 77.55\\
    LOL (full) & Oracle & VQA+Ours & 86.55 & 53.42 & 61.58 &71.04 & & 85.79 & 88.51 & 87.15\\
    LOL (full) & Trained & VQA+Ours & 86.55 & 53.42 & 61.58 &71.04 & & 82.13 & 84.17 & 83.15\\
    \midrule
    LXMERT & None & VQA+Ours & 85.23 & 51.25 & 60.58 & 69.78 & & 75.31 & 85.25 & 80.28 \\
    % LOL & None & VQA+Ours & 86.55 & 53.42 & 61.58 & 71.04 & & 72.88 & 88.32 \\
    LOL ($\mathit{q}$ATT) & None & VQA+Ours &  86.79 & 52.66 & 61.85 & 71.19 & & 79.88 & 87.12 & 83.50\\
    LOL (full)
    % ($\mathit{q}$ATT+$\ell$ATT) 
    & None & VQA+Ours & 86.55 & 53.42 & 61.58 &71.04 & & \underline{\textbf{82.39}} & \underline{\textbf{87.80}} & 85.10\\
    \bottomrule
    \end{tabular}
    }
    \end{center}
    
    \label{table:sota}
\end{table*}
    \subsection{Evaluation on VQA v2.0 Test Data}
    Table~\ref{table:sota} shows the performance the VQA Test-Standard datset.
    Our models maintain overall performance on the VQA test dataset, and at the same time substantially improve from random performance ($\sim$ 50\%) on logically composed questions to 82.39\% on \texttt{VQA-Compose} and 87.80\% on \texttt{VQA-Supplement}.
    This shows that logical connectives in questions can be learned while not degrading the overall performance on the original VQA test set (our models are within $\sim$1.5\% of the state-of-the-art on all three types of questions on the VQA test-set).

%%%%%%%%%%%%%%%%%%%%%%%%%%%%%%%%%%%%%%%%%%%%%%%%%%%%%%%%%%%%%%%%%%%%%%%%%%%%%%%%
\section{Discussion}
%%%%%%%%%%%%%%%%%%%%%%%%%%%%%%%%%%%%%%%%%%%%%%%%%%%%%%%%%%%%%%%%%%%%%%%%%%%%%%%%
Consider the example, {\it ``Is every boy who is holding an apple or a banana, not wearing a hat?"}, humans are able to answer it to be true if and only if each boy who is holding \textit{at least one} of an apple or a banana is not wearing a hat~\cite{arlotti1263}.
Natural language contains such complex logical compositions, not to mention ambiguities and the influence of context.
In this paper, we focus on the simplest -- negation, conjunction, and disjunction.
We have shown that existing VQA models are not robust to questions composed with these logical connectives, even when we train parsers to split the question into its components.
When humans are faced with such questions, they may refrain from giving binary (Yes/No) answers.
For instance, logically, the question{\it ``Did you eat the pizza and did you like it?"} has a negative answer if either of the two component questions has a negative answer.
However, humans might answer the same question with the answer {\it ``Yes, but I did not like it"}.
While human question-answering is indeed elaborate, explanatory, and clarifying, that is the scope of our future work; here we focus only on predicting a single binary answer.

We have shown how connectives in a question can be identified by enhancing LXMERT encoders with dedicated attention modules and loss functions.
We would like to stress on the fact that we do not use knowledge of the connectives during inference, but instead train the network to be aware of it based on cross-modal features, instead of predicting purely based on language model embeddings which fail to capture these nuances.
Our work is an attempt to modularize the understanding of logical components to train the model to utilize the outputs of the attention modules.
We believe this work has potential implications on logic-guided data augmentation, logically robust question answering, and for conversational agents (with or without images).
Similar strategies and learning mechanisms may be used in the future to operate ``logically'' in the image-space at the level of object classes, attributes, or semantic segments.

\section{Conclusion}
In this work, we investigate VQA in terms of logical robustness.
The key hypothesis is that the ability to answer questions about an image, must be extendable to a logical composition of two such questions.
We show that state-of-the-art models trained on VQA dataset lack this.
Our solution involves the ``Lens of Logic" model architecture that learns to answer questions with negation, conjunction, and disjunction.
We provide \texttt{VQA-Compose} and \texttt{VQA-Supplement}, two datasets containing logically composed questions to serve as benchmarks.
Our models show improvements in terms of answering these questions, while at the same time retaining performance on the original VQA test-set.

\section*{Acknowledgments}
Support from NSF Robust Intelligence Program (1816039 and 1750082), DARPA (W911NF2020006) and ONR (N00014-20-1-2332) is gratefully acknowledged.


\clearpage
% updated April 2002 by Antje Endemann
% Based on CVPR 07 and LNCS, with modifications by DAF, AZ and elle, 2008 and AA, 2010, and CC, 2011; TT, 2014; AAS, 2016; AAS, 2020

%% orig

% INITIAL SUBMISSION - The following two lines are NOT commented
% CAMERA READY - Comment OUT the following two lines
% \usepackage{ruler}
% \usepackage[width=122mm,left=12mm,paperwidth=146mm,height=193mm,top=12mm,paperheight=217mm]{geometry}

% \begin{document}
% \linenumbers
% \pagestyle{headings}
% \mainmatter

\title{Supplementary Material } % Replace with your title

%******************


\titlerunning{Supplementary Material for VQA-LOL}

\author{
% Tejas Gokhale\thanks{Equal Contribution}\orcidID{0000-0002-5593-2804} \and
% Pratyay Banerjee \printfnsymbol{1}\orcidID{0000-0001-5634-410X}  \and
% Chitta Baral\orcidID{0000-0002-7549-723X} \and 
% Yezhou Yang\orcidID{0000-0003-0126-8976}
}

\authorrunning{
% T. Gokhale et al.
}
\institute{
% Arizona State University, United States\\
% \email{\{tgokhale, pbanerj6, chitta, yz.yang\}@asu.edu
% }
}

\maketitle

\begin{abstract}
In our paper, we investigated visual question answering (VQA) through the lens of logical transformation.
We showed that state-of-the-art VQA models are unable to reliably predict answers for questions composed with logical operations, i.e. negation, conjunction, and disjunction.
We introduced new datasets VQA-Compose and VQA-Supplement, created with logical composition and a novel methodology to train models to learn logical operators in questions
In this supplementary material, we elaborate upon the following topics:
\begin{itemize}[noitemsep]
    \item Data creation process,
    \item Dataset analysis,
    \item Training datasets used for each experiment,
    \item Additional details about model training and hyper-parameters,
    \item Additional details about parser models, and
    \item Further analysis and insights about our results.
\end{itemize}

\end{abstract}

%%%%%%%%%%%%%%%%%%%%%%%%%%%%%%%%%%%%%%%%%%%%%%%%%%%%%%%%%%%%%%%%%%%%%%%%%%%%%%%%%%%%%%%%%%%%%
\section{Dataset Creation}
%%%%%%%%%%%%%%%%%%%%%%%%%%%%%%%%%%%%%%%%%%%%%%%%%%%%%%%%%%%%%%%%%%%%%%%%%%%%%%%%%%%%%%%%%%%%%
\begin{table}[t]
    \centering
    \caption{Examples of question negation. $Q$ denotes the original question from the VQA dataset, $\neg Q$ denotes its negation.}
    \begin{tabular}{p{0.5\linewidth} p{0.5\linewidth}}
        \toprule
        \textbf{$Q$} & \textbf{$\neg Q$} \\
        \toprule 
        Is this an area near the city ? & Is an this area \textit{not} near the city?\\
        Are all the men wearing ties ? & Are all the men \textit{not} wearing ties?\\
        Is there a chair ? & Is there \textit{no} chair?\\
        Do you think it's gonna rain? & Do you think it's \textit{not} gonna rain?\\
        \bottomrule
    \end{tabular}
    \label{tab:Q_neg}
\end{table}

The key idea behind our dataset creation process is to leverage existing annotations from the VQA-v2 dataset~\cite{antol2015vqa} and from MS-COCO~\cite{lin2014microsoft} which is the source of images in VQA-v2.
We use questions from VQA-v2, and object annotations and captions from MS-COCO for each image.

In order to create logically composed questions, we first filter out the ``yes-no" questions which constitute $38\%$ of the VQA dataset.
We further filter these by retaining only those yes-no questions with a single valid answer.
These questions which are $20\%$ of the VQA data, have an unambiguous answer, chosen unanimously by all human annotators who created the VQA dataset.
This satisfies the definition of \textit{``closed questions"}~\cite{bobrow1964natural} that we use, and are thus the atoms of our data creation process.

We use two closed questions corresponding to the same image to create logically composed questions using the Boolean operators: negation ($\neg$), conjunction ($\wedge$), and disjunction ($\vee$).
Since they have a clear unambiguous answer that is either ``yes" or ``no", we can treat them as Boolean variables, and obtain answers for every new question composed.
For negating a question, we follow a template-based procedure negates the question by adding a ``no" or ``not" before a verb, preposition or noun phrase, as shown in Table~\ref{tab:Q_neg}.
Note that our data creation method chooses to put a “‘not” or “no” either before a preposition, verb, or noun phrase. 
For instance, 
\textit{ Is this an area near the city?}
is transformed to either \textit{ Is this not an area near the city?} or \textit{Is this an area not near the city?} randomly.
Conjunction and disjunction are straightforward, we add the words ``and" and ``or" between two closed questions.

    \begin{figure}[t]
        \centering
        \includegraphics[width=\linewidth]{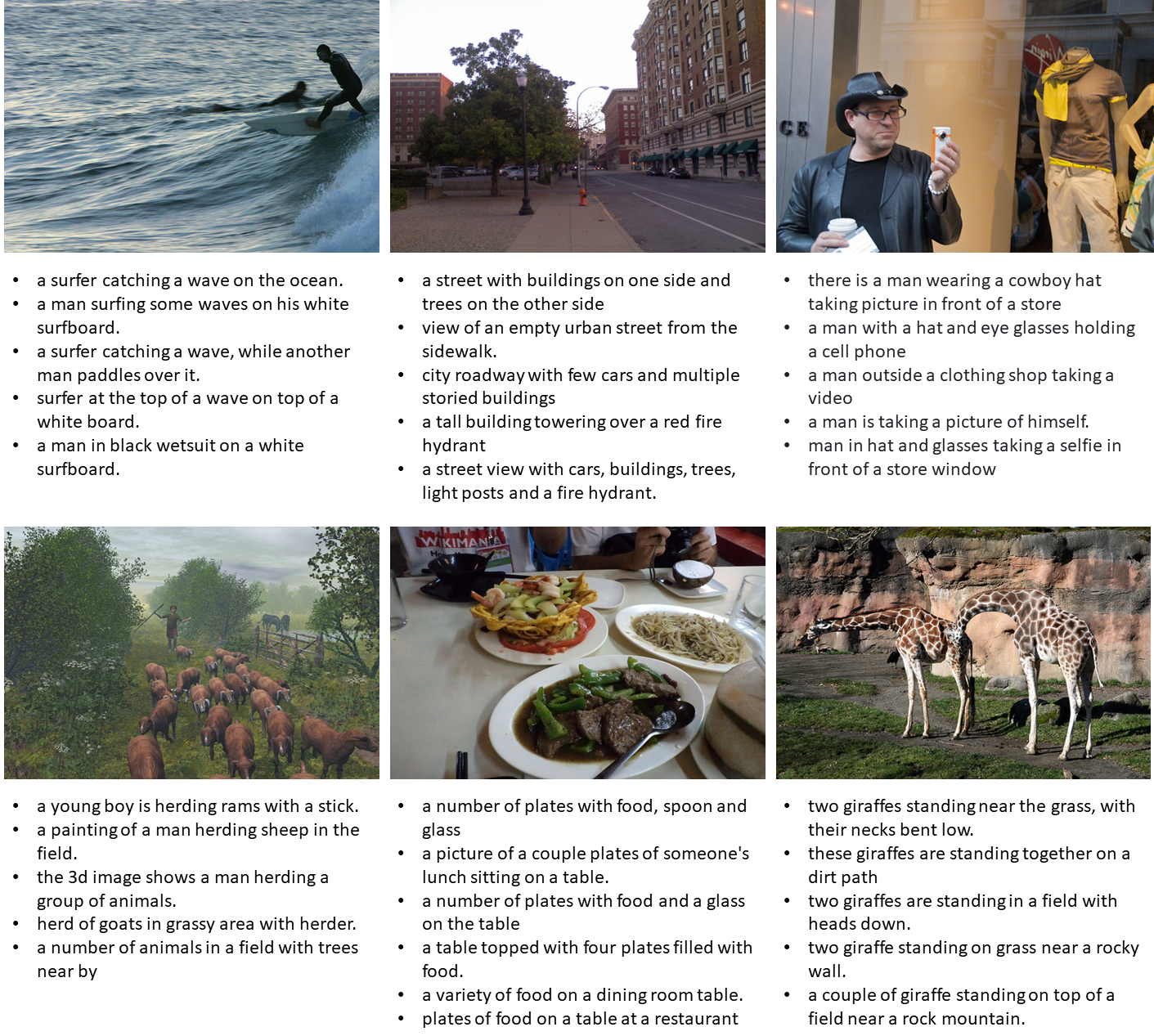}
        \caption{Examples of captions from COCO for images in the VQA dataset. 
        We convert these captions into questions and use them for our VQA-Supplement dataset}
        \label{fig:coco_captions}
    \end{figure}
    \begin{figure*}[!h]
        \centering
        \includegraphics[width=\linewidth]{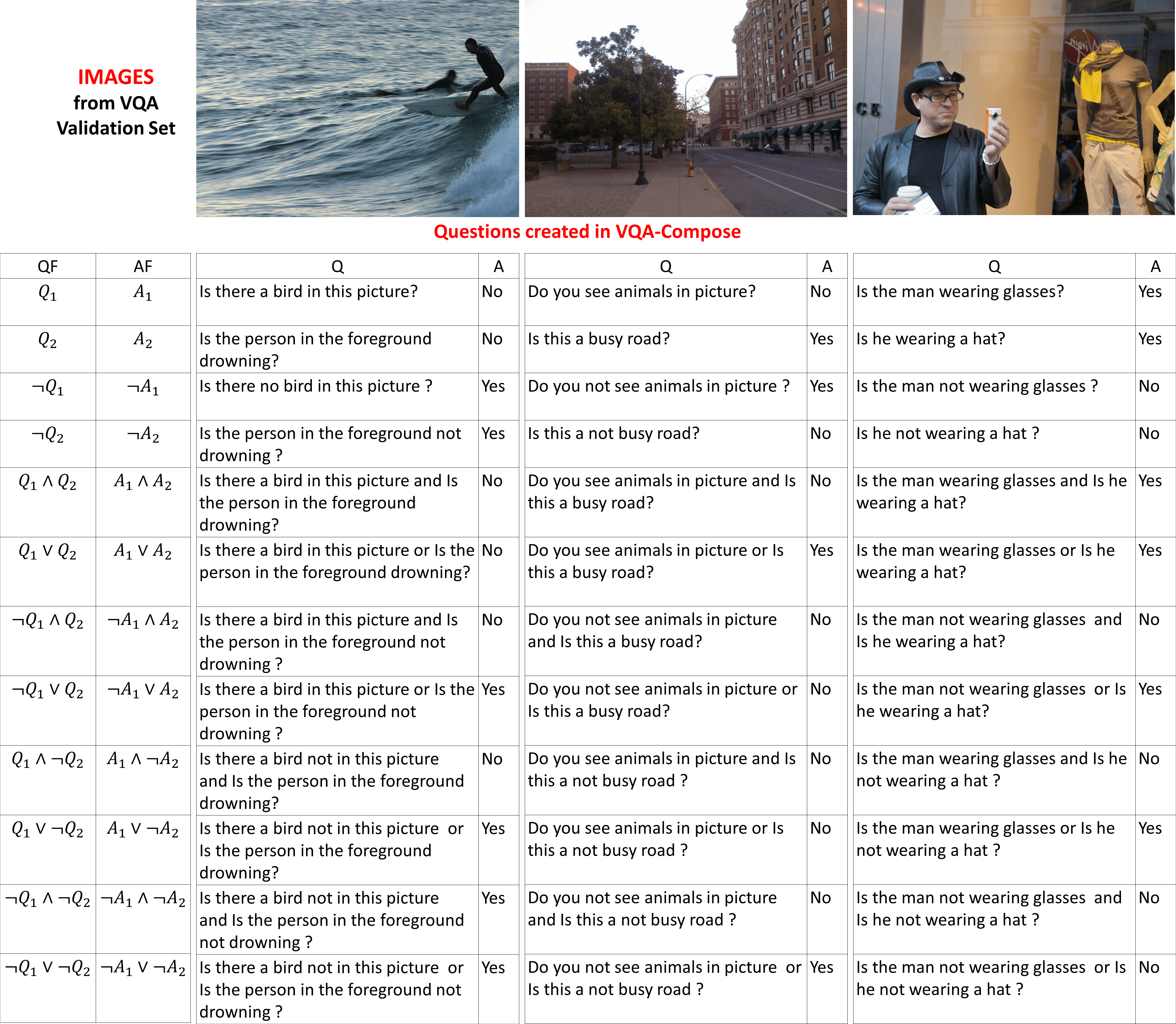}
        \caption{Some examples from our \texttt{VQA-Compose dataset}. 
        We show all 10 types of new questions created by original questions $Q_1$ and $Q_2$ and the corresponding answers. 
        Q, A, QF, AF denote question, answer, question-formula, and answer-formula respectively.
        anto(B) represents the adversarial antonym of objects in present in the image.
        }
        \label{fig:examples_lol}
    \end{figure*}
    
    \subsection{VQA-Compose}
    \texttt{VQA-Compose} is our dataset that is created solely from closed questions in the VQA dataset, by using negation, conjunction and disjunction to compose questions.
    As shown in Figure~\ref{fig:examples_lol}, we obtain 10 questions for each closed question in the VQA dataset, resulting in a total of 1.25M question-answer-image triplets as our \texttt{VQA-Compose} dataset.
    
    \subsection{VQA-Supplement}
    \begin{table}[t]
    \centering
    \caption{Examples of adversarial antonyms for objects.
    The antonym is chosen such that it is not in the image, but is semantically close to an object in the image}
    \begin{tabular}{lll} %{l cc cc cc cc cc cc cc c}
        \toprule
    %   \textbf{ Object} & bottle & \hphantom & cup & \hphantom  & spoon & \hphantom  & surfboard & \hphantom  & skis & \hphantom  & motorcycle & \hphantom  & sink & \hphantom  & suitcase   \\
    %   \textbf{Adv. Antonym}  & wine glass && bowl  && fork && skateboard && surfboard && bicycle && toilet && backpack \\
      \textbf{ Object} & \phantom{abcde} & \textbf{Adversarial Antonym} \\
      \toprule 
        bottle  && wine glass    \\
        cup     && bowl          \\
        spoon   && fork          \\
        surfboard && skateboard  \\
        % skis    &&   snowboard \\
        motorcycle && bicycle \\
        sink && toilet \\
        % suitcase && backpack \\
         \bottomrule
    \end{tabular}
    \label{tab:antonyms}
\end{table}
    \begin{figure*}[!h]
        \centering
        \includegraphics[width=\linewidth]{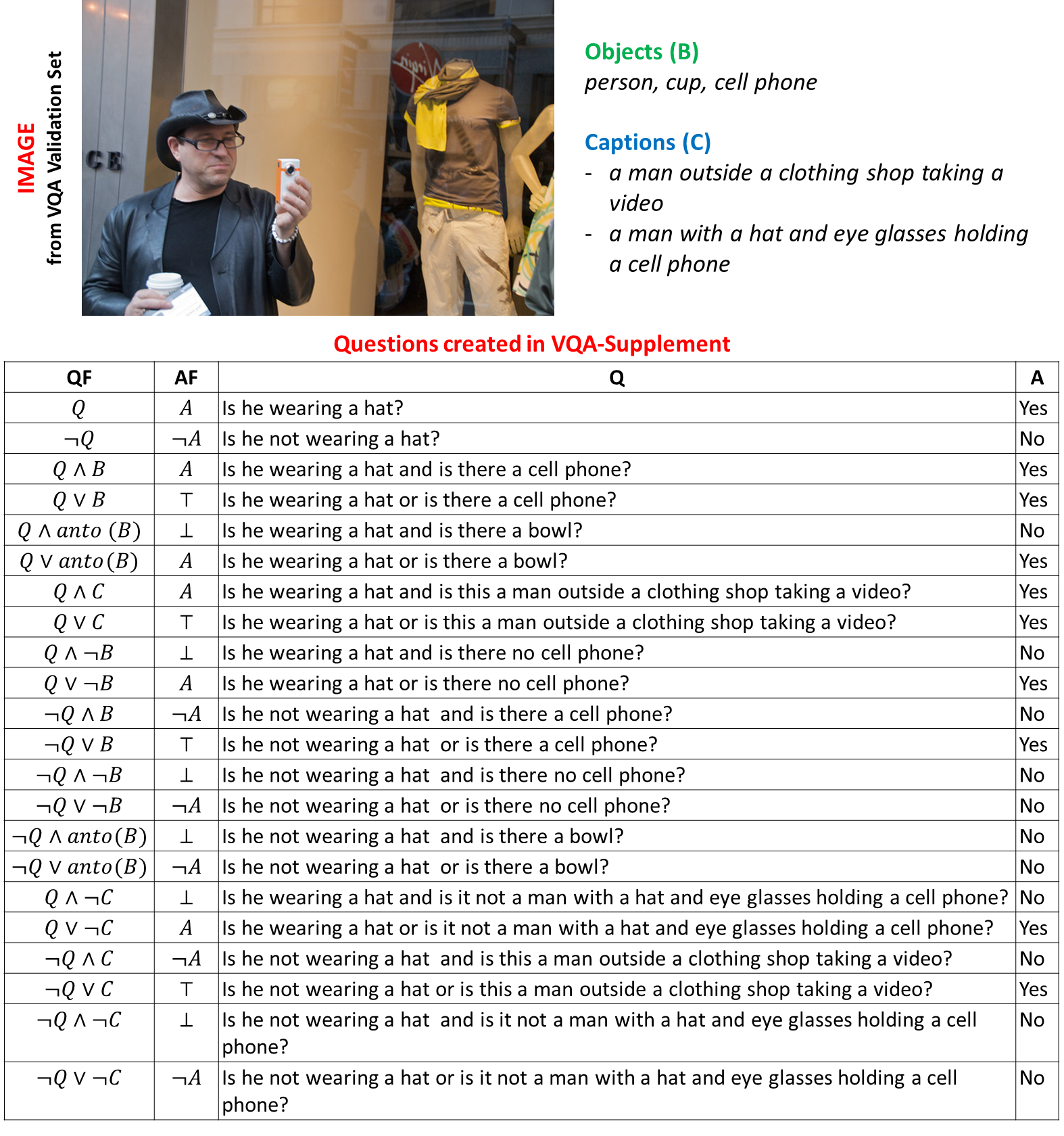}
        \caption{Some examples from our \texttt{VQA-Supplement dataset}. 
        We show all 20 types of new questions created by original questions $Q_1$ and $Q_2$ and the corresponding answers. 
        Q, A, QF, AF denote question, answer, question-formula, and answer-formula respectively.
        $\top, \bot$ are the standard Boolean symbols for top and bottom (true and false) }
        \label{fig:examples_cwl}
    \end{figure*}

    Figure~\ref{fig:coco_captions} shows examples of captions available in the MS-COCO dataset for images in the VQA-v2 dataset.
    As shown in Figure~\ref{fig:examples_cwl}, we use object annotations and captions from MS-COCO to create questions $B$ and $C$ respectively, using template-based methods.
    We create \texttt{VQA-Supplement} by using logical operators (negation, conjunction, and disjunction) to combine $B$ or $C$ with original questions from VQA-v2.
    
    In addition, we generate questions about adversarial object antonyms.
    An \textit{adversarial object antonym} is defined as an object that is not present in the image, but is closest semantically to an object in the image.
    Examples are shown in Table~\ref{tab:antonyms}.
    We use Glove vectors~\cite{pennington2014glove} to obtain embeddings of all object class names in the COCO dataset.
    Then for each image, we find adversarial antonyms using these vectors by using $\ell_2$ distance as a metric to sort and select adversarial antonyms.
    Since the list of objects present in the image is available to us via MS-COCO, we are able to determine the ground-truth answers for object-based questions.

    For each question $Q$ we obtain 20 new object-based and caption-based questions.
    In total, our \texttt{VQA-Supplement} dataset contains 2.55M question-answer-image triplets.

% %%%%%%%%%%%%%%%%%%%%%%%%%%%%%%%%%%%%%%%%%%%%%%%%%%%%%%%%%%%%%%%%%%%%%%%%%%%%%%%%%%%%%%%%%%%%
\section{Dataset Analysis}
% %%%%%%%%%%%%%%%%%%%%%%%%%%%%%%%%%%%%%%%%%%%%%%%%%%%%%%%%%%%%%%%%%%%%%%%%%%%%%%%%%%%%%%%%%%%%
In this section, we analyze the VQA dataset as well as our new datasets that contain logically composed questions.

\subsection{Question Length}
The average length of questions in VQA-v2~\cite{antol2015vqa} is \textbf{6.1 words}.
Our datasets have a average length of \textbf{12.25 words} for \texttt{VQA-Compose} and \textbf{15.17} for \texttt{VQA-Supplement}.
This is longer than VQA-v2 since each of our logically composed questions is made up of multiple component questions.

\begin{table}[!h]
    \centering
    \caption{Selected results of k-means clustering on the Glove embeddings of answers in VQA. k=50.}
    \resizebox{\linewidth}{!}{
    \begin{tabular}{lcl}
        \textbf{Cluster Name} &  & \textbf{Cluster Members} \\
        \toprule
    
        \pbox{0.11\linewidth}{Food} && \pbox{\linewidth}{'cooking', 'fast food', 'dishes', 'serving', 'grill', 'pizza hut', 'pizza box', 'lunch', 'restaurant', 'cafe', 'dinner', 'dairy', 'deli', 'menu', 'breakfast', 'cat food', 'burrito', 'food', 'dog food', 'eaten', 'burger', 'french fries', 'food processor', 'pizza cutter', 'grocery store', 'chef', 'pizza', 'vegetarian', 'eat', 'cook', 'food truck', 'chips', 'burgers', 'grocery', 'on pizza', 'eating', 'bar', 'sushi', 'sandwich', 'sandwiches', 'bars'}\\
        \midrule
        \pbox{0.11\linewidth}{Geography, Language, Ethnicity} && \pbox{\linewidth}{'china', 'thailand', 'america', 'american', 'africa', 'mexican', 'indians', 'russian', 'arabic', 'caucasian', 'american flag', 'german', 'russia', 'oriental', 'japan', 'hispanic', 'british', 'american airlines', 'asian', 'african american', 'italian', 'virgin', 'chinese', 'spanish', 'india', 'thai', 'japanese', 'asia', 'brazil', 'french', 'african', 'persian', 'english'}\\
        \midrule
        \pbox{0.11\linewidth}{Flowers, Plants} && \pbox{\linewidth}{'tulip', 'weeds', 'windowsill', 'tree branch', 'daffodils', 'carnations', 'elm', 'fern', 'grass', 'roses', 'garden', 'wreath', 'trees', 'pine', 'carnation', 'evergreen', 'sunflowers', 'tree', 'palm tree', 'ivy', 'palm', 'lily', 'iris', 'willow', 'christmas tree', 'vase','bamboo', 'tulips', 'rose', 'bushes', 'lilac',  'dandelions', 'plant', 'orchid', 'flowers', 'lilies', 'vines', 'daisy', 'cactus', 'palm trees', 'flower', 'floral', 'branches', 'bark', 'maple leaf', 'leaf', 'daffodil'}\\
        \midrule
       \pbox{0.11\linewidth}{Fruits} && \pbox{\linewidth}{'mango', 'apples', 'juice', 'cherries', 'strawberries', 'ginger', 'watermelon', 'cane', 'cherry', 'sweet', 'peach', 'organic', 'cantaloupe', 'orange juice', 'banana split', 'ripe', 'lemonade', 'grape', 'fruit', 'sunflower', 'smoothie', 'coconut', 'strawberry', 'banana peel', 'peaches', 'sesame seeds', 'fresh', \dots, 'mint', 'lemons', 'pineapple', 'oranges', 'grapes', 'salt and pepper', 'grapefruit', 'almonds', 'blueberry', 'kiwi'}\\
        \midrule
        \pbox{0.11\linewidth}{Birds} && \pbox{\linewidth}{'crows', 'pelicans', 'seagull', 'squirrel', 'finch', 'feathers', 'sparrow', 'stork', 'duck', 'parrots', 'rooster', 'eagle', 'bird feeder', 'peacock', 'bird', 'birds', 'goose', 'pigeon', 'crow', 'pigeons', 'owl', 'hummingbird', 'feeder', 'hawk', 'cranes', 'geese', 'flamingo', 'cardinal', 'nest', 'swan', 'ducks', 'parakeet', 'seagulls', 'parrot', 'woodpecker', 'swans', 'pelican'}\\
        \midrule
        \pbox{0.11\linewidth}{Sports} && \pbox{\linewidth}{'tennis shoes', 'playing game', 'playing baseball', 'tennis', 'baseball bat', 'tennis court', 'football', 'soccer', 'playing video game', 'sports', 'tennis racket', 'baseball uniform', 'team', 'bowling', 'hockey', 'play', 'baseball glove', 'goalie', 'playing tennis', 'badminton', 'playing frisbee', 'tennis player', 'rugby', 'soccer field', 'play tennis', 'soccer ball', 'athletics', 'basketball', \dots}\\
        \midrule
        \pbox{0.11\linewidth}{Dog Breeds} && \pbox{\linewidth}{'puppy', 'mutt', 'pomeranian', 'dogs', 'dachshund', 'bulldog', 'cocker spaniel', 'schnauzer', 'rottweiler', 'pitbull', 'pug', 'corgi', 'golden retriever', 'german shepherd', 'clydesdale', 'greyhound', 'boxer', 'kitten', 'cat', 'chihuahua', 'dog', 'husky', 'leash', 'terrier', 'dalmatian', 'thoroughbred', 'shepherd', 'sheepdog', 'collie', 'poodle', 'tabby', 'labrador', 'meow', 'beagle', 'calico', 'shih tzu', 'siamese'}\\
        \midrule
        \pbox{0.11\linewidth}{Colors} && \pbox{\linewidth}{'yellow and red', 'white and blue', 'green and red', 'neon', 'red bull', 'silver and red', 'blue', 'opaque', 'pink and blue', 'orange and yellow', 'black and brown', 'gray and white', 'brown and white', 'blue and black', 'maroon', 'yellow', 'silver', 'gray and red', 'orange and black', 'white and brown', 'black and red', 'black and yellow', 'green', 'purple',  'red and silver', 'colored', 'white and gray', 'black and gray'}\\
        \midrule
        \pbox{0.11\linewidth}{Sports Teams} && \pbox{\linewidth}{'dodgers', 'mariners', 'mets', 'cardinals', 'braves', 'yankees', 'phillies', 'orioles'}\\
        \midrule
        \pbox{0.11\linewidth}{Vegetables} && \pbox{\linewidth}{'cauliflower', 'sliced', 'lettuce', 'celery', 'parsley', 'basil', 'squash', 'peppers', 'beets', 'sesame', 'cucumber', 'onion', 'asparagus', 'carrots', 'mushrooms', 'mustard', 'beans', 'broccoli and carrots', 'carrot', 'cilantro', 'cabbage', 'tomato', 'feta', 'veggies', 'avocado', 'peas', 'garlic', 'zucchini', 'pepper', 'vegetables', 'potatoes', 'tomatoes', 'radish', }\\
        \midrule
        \pbox{0.11\linewidth}{Bathroom} && \pbox{\linewidth}{'toothbrushes', 'lotion', 'washing', 'toiletries', 'faucet', 'mouthwash', 'towel', 'urinal', 'above toilet', 'toothpaste', 'soap', 'pooping', 'bathtub', 'bathing', 'tub', 'drain', 'toilet brush', 'pee', 'shampoo', 'towels', 'on toilet', 'shower', 'bidet', 'toilet paper', 'peeing', 'laundry', 'toilets', 'shower head', \dots}\\
        \midrule
        \pbox{0.11\linewidth}{Clothes} && \pbox{\linewidth}{'life jacket', 'hat', 'fabric', 'shirts', 'apron', 'bathing suit', 'adidas', 'belt', 'pocket', 'sweater', 't shirt', 'slacks', 'jeans', 'zipper', 'vests', 'bandana', 'costume', 'jackets', 'hoodie', 'strap', 'jacket', 'shoes', 'bow tie', 'pockets', 'yarn', 'denim', 'socks', 't shirt and jeans', 'khaki', 'tuxedo', 'shirt', 'robe', 'swimsuit', 'sleeve', 'overalls', 'uniform', 'cap', 'clothing', 'camouflage', 'fedora', 'suits', 'boots', \dots}\\
        \bottomrule
    \end{tabular}
    }
    \label{tab:clusters}
\end{table}
\subsection{Types of Answers}
The VQA dataset contains a fixed vocabulary of answers.
We obtained the Glove~\cite{pennington2014glove} embeddings of these answers, and performed k-means clustering on these embeddings to obtain 50 clusters.
We show examples of some of these clusters in Table~\ref{tab:clusters}.
It can be observed that similar answers, such as those belonging a common category such as \textit{food} or \textit{sports} appear in the same cluster.
This shows that Glove embeddings of these answers preserve a notion of similarity.
Note that the cluster names in Table~\ref{tab:clusters} are assigned by humans after clustering is complete, for the sake of clarity and illustration, and does not play a role in the clustering process.
It is interesting to know that our cluster categories are similar to ``knowledge categories" obtained in OK-VQA ~\cite{marino2019ok}.
The categories in OK-VQA are annotated by human workers in Amazon Mechanical Turk.

\clearpage

%%%%%%%%%%%%%%%%%%%%%%%%%%%%%%%%%%%%%%%%%%%%%%%%%%%%%%%%%%%%%%%%%%%%%%%%%%%%%%%%%%%%%%%%%%%%%
\section{Training Data for Our Experiments}
%%%%%%%%%%%%%%%%%%%%%%%%%%%%%%%%%%%%%%%%%%%%%%%%%%%%%%%%%%%%%%%%%%%%%%%%%%%%%%%%%%%%%%%%%%%%%

For each experimental setting, we train our models with a dataset containing questions from VQA, \texttt{VQA-Compose}, and \texttt{VQA-Supplement}.
The proportions of these samples in the training data depends upon the specific experiment performed.
For each of our experiments we use the same train-validation-test splits as in the VQA-v2 and COCO datasets.
In this section, we explain our training datasets in detail for each experiment, analysis, and ablation study.

\begin{table}[t]
    \caption{Training dataset distribution and sizes, for explicit training with new data. Note that training dataset sizes are consistent with the VQA dataset.}
    \begin{center}
    \resizebox{\linewidth}{!}{
    \begin{tabular}{lccccccc}
    \toprule
    \multirow{2}{*}{\textbf{\pbox{15mm}{Training\\Datasets}}} & \multicolumn{5}{c}{\textbf{Proportion of datasets (\%)}} & \hphantom &  \multirow{2}{*}{\textbf{\pbox{25mm}{Training\\Samples}}}\\
    \cmidrule{2-6}
     & VQA-Other & VQA-Number & VQA-YesNo & \texttt{Comp} & \texttt{Supp} & \\
    \toprule
    VQA & 50 & 12 & 38 & 0 & 0 && 443754\\
    VQA+Comp &  50 & 12 & 19 & 19 & 0 && 443754\\
    VQA+Comp+Supp & 50 & 12 & 12.66 & 12.66 & 12.66 && 443754\\
    \bottomrule
    \end{tabular}
    }
    \end{center}
    
    \label{table:data_exp1}
\end{table}

\begin{table}[t]
    \caption{Training datasets distribution and sizes, for the experiment for understanding the effect of logically composed questions. We progressively add more logical samples, and get the learning curve as shown in the paper.}
    \begin{center}
    \resizebox{\linewidth}{!}{
    \begin{tabular}{lcccccc}
    \toprule
    \multirow{2}{*}{\textbf{{Training Datasets}}}
    & \multicolumn{5}{c}{\textbf{Proportion of samples (\%)}} & \multirow{2}{*}{\textbf{\pbox{20mm}{Training\\Samples}}}\\
    \cmidrule{2-6}
    & VQA-Other & VQA-Number & VQA-YesNo & \texttt{Comp} & \texttt{Supp} & \\
    \toprule
    VQA & 50 & 12 & 38 & 0 & 0 & 443754 \\
    \midrule
    VQA + Comp (10) & 49.999 & 11.999 & 37.999 & 0.002 & 0 & 443764\\
    VQA + Comp (100) & 49.989 & 11.997 & 37.991 & 0.022 & 0 & 443854\\
    VQA + Comp (1k) & 49.888 & 11.973 &  37.914 & 0.225 & 0 & 444754\\
    VQA + Comp (10k) & 48.898 & 11.736 &  37.162 & 2.204 & 0 & 453754\\
    VQA + Comp (100k) & 40.805 & 9.793 & 31.011 & 18.391 & 0 & 543754\\
    \midrule
    VQA + Comp (10) + Supp (10) & 49.998 & 11.999 &	37.998 & 0.002 & 0.002	 & 443774\\
    VQA + Comp (100) + Supp (100) & 49.977	& 11.995 & 37.983 & 0.022 & 0.022 & 443954\\
    VQA + Comp (1k)+ Supp (1k) & 49.776 & 11.946 & 37.829 &	0.224 & 0.224 & 445754\\
    VQA + Comp (10k)+ Supp (10k) & 47.844 &	11.483 & 36.361 & 2.156 & 2.156 & 463754\\
    VQA + Comp (100k)+ Supp (100k)& 34.466 &	8.272 & 26.194 & 15.534	& 15.534 & 643754\\
    \midrule 
    \end{tabular}
    }
    \end{center}
    
    \label{table:data_exp3}
\end{table}

\begin{table}
    \caption{Training datasets distribution and sizes, for training with logical questions with a maximum of one connective. 
    % Sizes of these datasets are the same as those in Table~\ref{table:exp2}
    }
    \begin{center}
    \resizebox{\linewidth}{!}{
    \begin{tabular}{lcccccc}
    \toprule
    \multirow{2}{*}{\textbf{{Training Datasets}}}
    & \multicolumn{5}{c}{\textbf{Proportion of samples (\%)}} & \multirow{2}{*}{\textbf{\pbox{20mm}{Training\\Samples}}}\\
    \cmidrule{2-6}
    & VQA-Other & VQA-Number & VQA-YesNo & \texttt{Comp-Single} & \texttt{Supp-Single} & \\
    \toprule
    YesNo &  0 & 0 & 100 & 0 & 0 & 168626\\
    YesNo + Comp &  0 & 0 & 50 & 50 & 0 & 337253 \\
    YesNo + Comp + Supp & 0 & 0 & 33.33 & 33.33 & 33.33 & 505879\\
    \bottomrule
    \end{tabular}
    }
    \end{center}
    
    \label{table:data_exp4}
\end{table}

\subsection{Explicit Training with new data}
In this experiment, we investigate if existing models trained on VQA data are able to answer questions in \texttt{VQA-Compose} and \texttt{VQA-Supplement}.
We compare this with the LXMERT model~\cite{tan2019lxmert} trained explicitly with our new data, and also with our models that use the attention modules for question-type and connective-type.
For a fair comparison, we restrict the size of training dataset to the original size of the VQA training dataset ($443,754$ samples).
We also use the same proportion of question-types as in VQA ($38\%$ yes-no, $12\%$ number, and $50\%$ other questions), as shown in Table~\ref{table:data_exp1}.
This allows us to improve the diversity of yes-no questions, by incorporating yes-no questions from \texttt{VQA-Compose} and \texttt{VQA-Supplement}.

\subsection{Training with Closed Questions only}
For this experiment, we evaluate the models when trained only on closed questions, under three settings: 
\begin{enumerate}[nosep,noitemsep]
    \item yes-no questions from VQA
    \item yes-no questions from VQA along with an equal number of questions from \texttt{VQA-Compose}, 
    \item yes-no questions from VQA along with an equal number of questions from \texttt{VQA-Compose} and \texttt{VQA-Supplement}
\end{enumerate}
This allows us to compare the capability of models to answer different types of yes-no questions  such as the original questions from VQA, logical compositions in \texttt{VQA-Compose}, and logical compositions with object and caption-based questions in \texttt{VQA-Supplement}.

\subsection{Effect of Logically Composed Questions}
In this experiment, we progressively add logically composed questions to the training data, and analyze the learning curve with respect to the number of logical samples
We add $10$, $100$, $1k$, $10k$, and $100k$ samples from \texttt{VQA-Compose} or both \texttt{VQA-Compose} and \texttt{VQA-Supplement}.
The training set distribution in shown in Table~\ref{table:data_exp3}.
This allows us to understand how many additional logically composed questions are needed for our models to become robust.

\subsection{Compositional Generalization}
In this experiment, our aim is to train models on questions that contain a single logical connective ({\it and, or, not}) or no connective at all (original yes-no questions in VQA), and to test their performance on questions with more than one connective.
To do so, we restrict our training data to such single-connective questions as shown in Table~\ref{table:data_exp4}

%%%%%%%%%%%%%%%%%%%%%%%%%%%%%%%%%%%%%%%%%%%%%%%%%%%%%%%%%%%%%%%%%%%%%%%%%%%%%%%%%%%%%
\section{Model Architectures and Training Settings}
%%%%%%%%%%%%%%%%%%%%%%%%%%%%%%%%%%%%%%%%%%%%%%%%%%%%%%%%%%%%%%%%%%%%%%%%%%%%%%%%%%%%%
\begin{table}[t]
    \centering
    \caption{Hyper-Parameters for training LXMERT and our models}
    \begin{tabular}{lcl}
        \toprule
        \textbf{Hyper-Parameters } & \hphantom & \textbf{Model} \\ 
        \toprule
        Batch Size             && 32                         \\ 
        Learning Rate          && 5e-5                       \\ 
        Dropout                && 0.1                        \\ 
        Language Layers        && 9                          \\ 
        Cross-Modality Layer   && 5                          \\ 
        Object Relation Layers && 5                          \\ 
        Optimizer              && BertAdam                   \\ 
        Warmup                 && 0.1                        \\ 
        Max Gradient Norm      && 5.0                        \\ 
        Max Text Length    && 20                         \\ 
        \bottomrule
    \end{tabular}
    \label{tab:hyper}
\end{table}
We train our models and baseline LXMERT~\cite{tan2019lxmert} model with the hyper-parameters in Table \ref{tab:hyper}, chosen from the median of 5 random seeds. 
The length of cross-modal embeddings produced by LXMERT for each question-image pair is 768.
We utilize this as input to our attention modules $\mathbf{q_{\textit{\tiny ATT}}}$ and $\mathbf{\ell_{\textit{\tiny ATT}}}$.
The hidden layers of these attention modules have a size of $2\times768$.
The answering module uses the outputs of these modules to predict softmax answer probabilities.

%%%%%%%%%%%%%%%%%%%%%%%%%%%%%%%%
\begin{table}[t]
\caption{Precision-Recall and F1-Scores for the RoBERTa-based NER parser}
\begin{center}
\begin{tabular}{p{2cm}p{2cm}p{2cm}p{2cm}}
\toprule
\textbf{Operands} & \textbf{Precision} & \textbf{Recall} & \textbf{F1-Score }\\
\toprule
% \hline
2 & 84.98 & 86.69 & 85.83 \\
% \hline
3 & 81.55 & 83.62 & 82.57 \\
% \hline
4 & 81.63 & 83.72 & 82.66 \\
% \hline
5 & 76.29 & 79.45 & 77.84 \\
\bottomrule
\end{tabular}
\end{center}

\label{tab:parser}
\end{table}
\section{Parser Training and Results}
One of our baselines involves using a parser to split a question into its components, answer them separately, and combine the answers logically to get the final answer.
We use the RoBERTa-Base language model \cite{liu2019roberta} and train it for the Named-Entity Recognition (NER) task.  
We modify the RoBERTa-NER model from the Huggingface Transformers \cite{Wolf2019HuggingFacesTS} framework.  
We create our parser dataset using the constituent questions as target entities and the original question as the input text. The sequence is classified using B-I-O \textit{(Beginning-Inside-Outside)}~\cite{liu2019roberta} tagging scheme, where all constituent tokens are predicted to be tagged as B-Const, I-Const and the connectives are tagged as O. 
\footnote{``Const" refers to constituent.}
There is only one entity class. 
We train the model for $20$ epochs, with a batch size of $32$, and learning rate of $1e\text{-}5$.
The results of our parser are shown in Table~\ref{tab:parser}.
It can be observed that the performance of the parser deteriorates as the number of operands in the question increases.
This is a major drawback of parser-based methods.

%%%%%%%%%%%%%%%%%%%%%%%%%%%%%%%%%%%%%%%%%%%%%%%%%%%%%%%%%%%%%%%%%%%%%%%%%%%%%%%%%%%%%%%%%%%%%
\section{Analysis of Results}
%%%%%%%%%%%%%%%%%%%%%%%%%%%%%%%%%%%%%%%%%%%%%%%%%%%%%%%%%%%%%%%%%%%%%%%%%%%%%%%%%%%%%%%%%%%%%
\begin{figure}[t]
    \begin{center}
        \subfloat[\label{supp_heatmap_comp}]{%
        \includegraphics[angle=0,origin=c,width=0.49\linewidth]{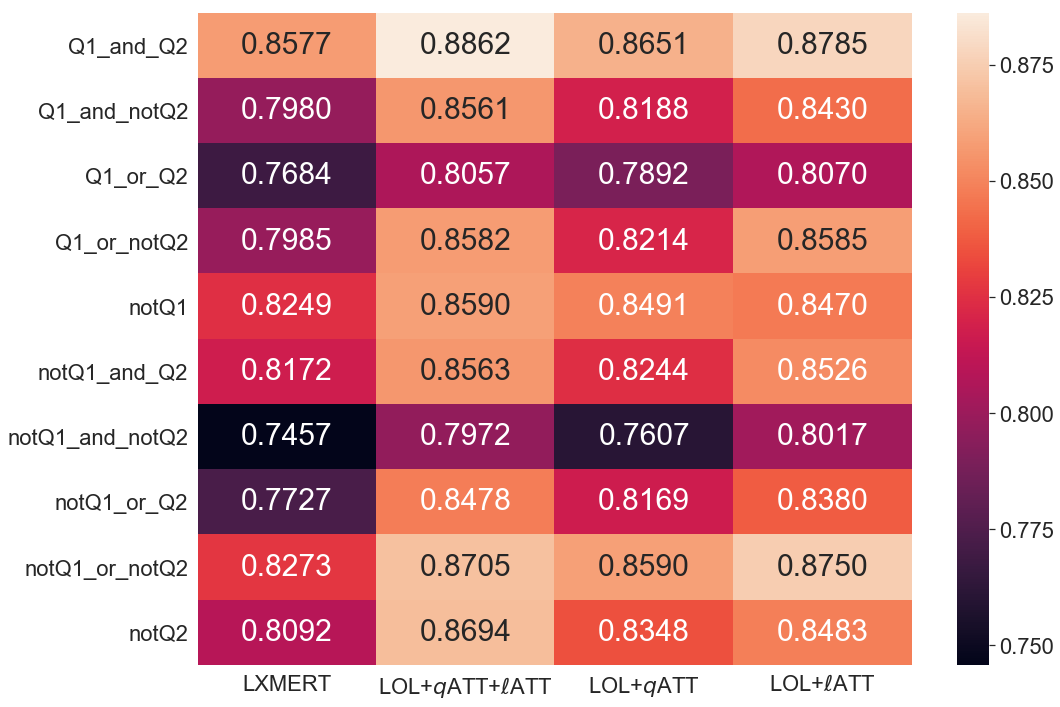}
        }
        \subfloat[\label{supp_heatmap_supp}]{%
        \includegraphics[angle=0,origin=c,width=0.49\linewidth]{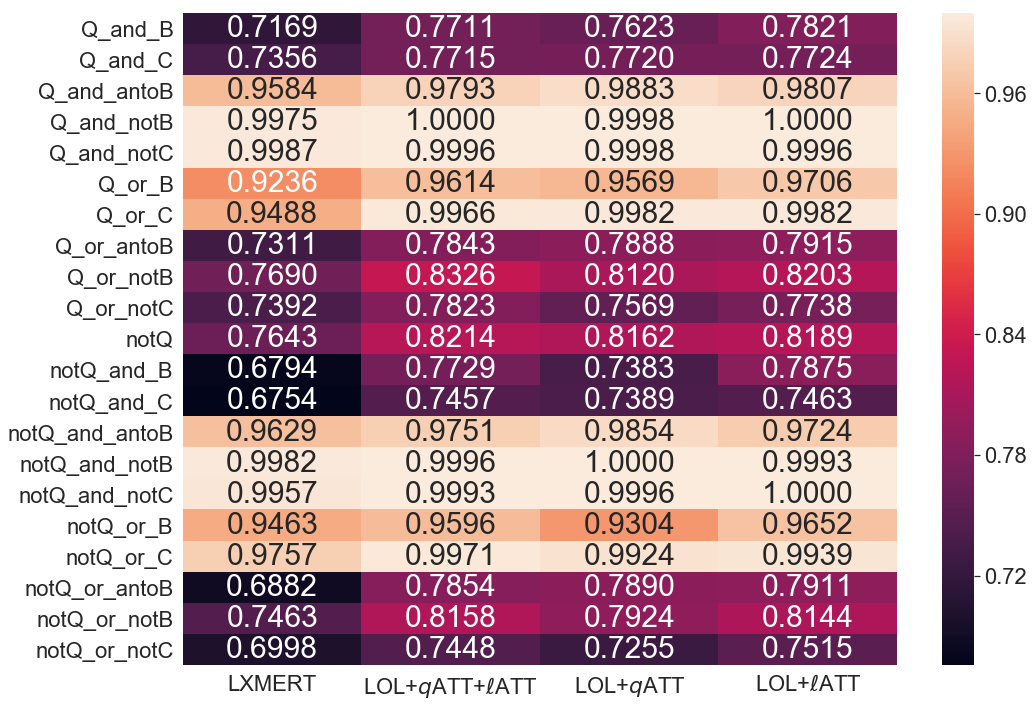}
        }
    \end{center}
    \caption{Accuracy for each type of question in (a) VQA-Compose, (b) VQA-Supplement and for questions with number of operands greater than 2.}
    \label{fig:heatmap_both}
\end{figure}
\begin{table}[t]
    \caption{Accuracies on each type of question in \texttt{VQA-Compose} by each model. QF is Question Formula}
    \begin{center}
    \resizebox{\linewidth}{!}{
        \begin{tabular}{lcc cc cc c}
    \toprule
    \textbf{QF} & \textbf{LXMERT} &\hphantom& \textbf{LXMERT+$\ell_{ATT}$} &\hphantom & \textbf{LXMERT+$\mathbf{q_{\textit{\tiny ATT}}}$} &\hphantom & \textbf{LXMERT+$\mathbf{q_{\textit{\tiny ATT}}}$+$\mathbf{\ell_{\textit{\tiny ATT}}}$}\\
    \toprule
    $\neg Q_1$                  & 85.39 && 85.55 && 84.78 && 86.43\\
    $\neg Q_2$                  & 84.38 && 85.45 && 84.94 && 86.08\\
    $Q_1\wedge Q_2$             & 81.50 && 87.77 && 87.66 && 87.77\\
    $Q_1 \vee Q_2$              & 85.26 && 81.58 && 80.54 && 80.97\\
    $Q_1 \wedge \neg Q_2$       & 85.71 && 85.77 && 84.45 && 85.02\\
    $Q_1 \vee \neg Q_2$         & 87.12 && 86.22 && 85.98 && 85.53\\
    $\neg Q_1 \wedge Q_2$       & 85.10 && 85.34 && 84.83 && 85.53\\
    $\neg Q_1 \vee Q_2$         & 80.76 && 78.92 && 83.79 && 84.75\\
    $\neg Q_1 \wedge \neg Q_2$  & 87.98 && 86.59 && 79.77 && 81.32\\
    $\neg Q_1 \vee \neg Q_2$    & 87.12 && 85.42 && 87.42 && 87.74\\
    \bottomrule
    \end{tabular}
    }
    \end{center}
    \label{table:heatmap_table_1}
\end{table}
\begin{table}[!h]
    \caption{Accuracies on each type of question in \texttt{VQA-Supplement} by each model}
    \begin{center}
    \resizebox{\linewidth}{!}{
    \begin{tabular}{lcc cc cc c}
    \toprule
    \textbf{QF} & \textbf{LXMERT} &\hphantom& \textbf{LXMERT+$\ell_{ATT}$} &\hphantom & \textbf{LXMERT+$\mathbf{q_{\textit{\tiny ATT}}}$} &\hphantom & \textbf{LXMERT+$\mathbf{q_{\textit{\tiny ATT}}}$+$\mathbf{\ell_{\textit{\tiny ATT}}}$}\\
    \toprule
$Q$                     &  82.27  &&  82.3  &&  82.77 &&  82.34  \\
$Q \wedge B$            &  78.03  &&  77.92  &&  78.16 && 78.36  \\
$Q \vee B$              &  95.51  &&  96.79  &&  97.06  &&  96.74  \\
$Q \wedge anto(B)$      &  95.64  &&  97.55  &&  98.07  &&  96.72  \\
$Q \wedge C$            &  81.22  & & 82.07  &&  81.67  &&  81.67  \\
$Q \vee C$              &  99.84  & & 99.89  &&  99.84  &&  99.89  \\
$Q \wedge \neg B$       &  99.96  &&  99.93  &&  99.98  &&  99.89  \\
$Q \vee \neg B$         &  82.39  &&  82.54  &&  82.09  &&  81.69  \\
$\neg Q \vee B$         &  95.08  &&  96.52  &&  96.52  &&  95.51  \\
$\neg Q \wedge \neg B$  &  99.89  &&  99.84  &&  99.91  &&  99.75  \\
$\neg Q \wedge anto(B)$ &  94.86  &&  97.91  && 97.15  &&  97.42  \\
$Q \wedge \neg C$       &  99.91  &&  99.91  &&  99.98  &&  99.87  \\
$Q \vee \neg C$         &  82.45  &&  82.21  &&  82.3  &&  81.46  \\
$\neg Q \vee C$         &  99.80  &&  99.91  &&  99.75  &&  99.82  \\
$\neg Q \wedge \neg C$  &  99.84  &&  99.87  &&  99.89  &&  99.78  \\
$\neg Q$                &  80.30  &&  81.62  &&  81.78  &&  80.84  \\
$Q \vee anto(B)$        &  77.92  &&  77.83  &&  79.13  &&  78.43  \\
$\neg Q \wedge B$       &  76.27  &&  76.90  && 78.88  &&  77.31  \\
$\neg Q \vee \neg B$    &  79.73  &&  81.42  &&  81.49  &&  81.17  \\
$\neg Q \vee anto(B)$   &  75.62  &&  77.33  &&  79.22  &&  77.92  \\
$\neg Q \wedge C$       &  78.95  &&  81.26  &&  81.11  &&  80.18  \\
$\neg Q \vee \neg C$    &  79.87  &&  80.77  &&  81.51  &&  80.61  \\
\bottomrule
    \end{tabular}
    }
    \end{center}

    \label{table:heatmap_table_2}
\end{table}
We provide accuracies of all four models as a heat-map in Figure~\ref{fig:heatmap_both}, and also in Tables \ref{table:heatmap_table_1} and \ref{table:heatmap_table_2}.
We have two key observations.

In Figure~\ref{supp_heatmap_comp}, we observe that for all models, the two hardest question categories are $Q_1 \vee Q_2$ and $\neg Q_1\wedge\neg Q_2$, while the two easiest categories are $Q_1\wedge Q_2$ and $\neg Q_1 \vee \neg Q_2$.
Using DeMorgan's laws to rewrite these logical formulas, we see that the two hardest categories are:
$$\mathbf{Q_1 \vee Q_2}~~,~~\mathbf{\neg(Q_1 \vee Q_2)},$$ 
while the two easiest categories are:
$$\mathbf{Q_1 \wedge Q_2}~~,~~\mathbf{\neg(Q_1 \wedge Q_2)}.$$

Figure~\ref{supp_heatmap_supp} provides similar insights.
Note that since questions $B$ and $C$ are composed from factually valid statements (about objects in the image, or from valid caption describing a scene), the answers to these questions are always ``Yes".
Thus answers to any question that uses a disjunction (``or") to combine $B, C$ with another question, is always ``Yes".
Similarly answers to $\neg B, \neg C, anto(B)$ are always ``No".
Thus answers to any question that uses a conjunction (``and") to combine  $\neg B, \neg C, anto(B)$ with another question, is always ``No".
These question categories are $Q\vee B, Q\vee C, \neg Q\vee B, \neg Q\vee C$, and $Q\wedge\neg B, Q\wedge\neg C, Q\wedge anto(B), \neg Q\wedge\neg B, \neg Q\wedge\neg C$, and $\neg Q\wedge anto(B)$.

It is interesting to note that questions about adversarial objects are relatively harder to answer for any category and any model, than the questions about objects present in the image.
Thus we see that answering questions about objects in the image is much easier than other categories for each model.

Following a similar trend, we observe a difficulty in answering questions which use conjunction (``and") to combine $B, C$ with another question, or which use disjunction (``and") to combine  $\neg B, \neg C, anto(B)$ with another question.
This is because the answer to these questions changes according to the sample and depends on the answer to the question $Q$, and cannot be simply ``explained away".

% \bibliography{egbib}
% \end{document}

\clearpage

\bibliographystyle{splncs04}
\bibliography{egbib}
\end{document}